\pdfoutput=1
\documentclass[11pt,a4paper]{article}
\usepackage[hyperref]{emnlp2020}
\usepackage{times}
\usepackage{latexsym}

\usepackage{microtype}

\usepackage{natbib}
\usepackage{multirow}

\def\equationautorefname~#1\null{equation~(#1)\null}

\usepackage{etoolbox}
\makeatletter
\patchcmd{\hyper@makecurrent}{%
	\ifx\Hy@param\Hy@chapterstring
	\let\Hy@param\Hy@chapapp
	\fi
}{%
	\iftoggle{inappendix}{
		\@checkappendixparam{chapter}%
		\@checkappendixparam{section}%
		\@checkappendixparam{subsection}%
		\@checkappendixparam{subsubsection}%
		\@checkappendixparam{paragraph}%
		\@checkappendixparam{subparagraph}%
	}{}%
}{}{\errmessage{failed to patch}}

\newcommand*{\@checkappendixparam}[1]{%
	\def\@checkappendixparamtmp{#1}%
	\ifx\Hy@param\@checkappendixparamtmp
	\let\Hy@param\Hy@appendixstring
	\fi
}
\makeatletter

\newtoggle{inappendix}
\togglefalse{inappendix}

\apptocmd{\appendix}{\toggletrue{inappendix}}{}{\errmessage{failed to patch}}

\usepackage{amsmath}
\usepackage{amsfonts}
\usepackage{amssymb}
\usepackage{amsthm}
\theoremstyle{definition}
\newtheorem{definition}{Definition}
\newtheorem{example}[definition]{Example}
\newtheorem{task}[definition]{Task}
\usepackage{bm}
\allowdisplaybreaks

\usepackage{arydshln}
\newcommand{\lp}{\texttt{(}}
\newcommand{\rp}{\texttt{)}}
\newcommand{\lb}{\texttt{[}}
\newcommand{\rb}{\texttt{]}}
\newcommand{\ta}{\texttt{a}}
\newcommand{\tb}{\texttt{b}}
\newcommand{\tc}{\texttt{c}}
\newcommand{\td}{\texttt{d}}
\newcommand{\ttt}[1]{\texttt{#1}}

\DeclareMathOperator{\sign}{sign}

\aclfinalcopy 

\title{Evaluating Attribution Methods using White-Box LSTMs}

\author{Yiding Hao \\
  Yale University \\
  New Haven, CT, USA \\
  \texttt{yiding.hao@yale.edu}}

\date{}

\begin{document}
\maketitle
\begin{abstract}
	Interpretability methods for neural networks are difficult to evaluate because we do not understand the black-box models typically used to test them. This paper proposes a framework in which interpretability methods are evaluated using  manually constructed networks, which we call \textit{white-box networks}, whose behavior is understood \textit{a priori}. We evaluate five methods for producing attribution heatmaps by applying them to white-box LSTM classifiers for tasks based on formal languages. Although our white-box classifiers solve their tasks perfectly and transparently, we find that all five attribution methods fail to produce the expected model explanations.
\end{abstract}

\section{Introduction}

\textit{Attribution methods} are a family of interpretability techniques for individual neural network predictions that attempt to measure the importance of input features for determining the model's output. Given an input, an attribution method produces a vector of \textit{attribution} or \textit{relevance scores}, which is typically visualized as a heatmap that highlights portions of the input that contribute to model behavior. In the context of NLP, attribution scores are usually computed at the token level, so that each score represents the importance of a token within an input sequence. These heatmaps can be used to identify keywords upon which networks base their decisions \citep[\textit{inter alia}]{liVisualizingUnderstandingNeural2016,sundararajanAxiomaticAttributionDeep2017,arrasWhatRelevantText2017,arrasExplainingRecurrentNeural2017,murdochWordImportanceContextual2018}. 

One of the main challenges facing the evaluation of attribution methods is that it is difficult to assess the quality of a heatmap when the network in question is not understood in the first place. If a word is deemed relevant by an attribution method, we do not know whether the model actually considers that word relevant, or whether the attribution method has erroneously estimated its importance. Indeed, previous studies have argued that attribution methods are sensitive to features unrelated to model behavior in some cases (e.g., \citealp{kindermansReliabilitySaliencyMethods2019}), and altogether insensitive to model behavior in others \citep{adebayoSanityChecksSaliency2018}.

To tease the evaluation of attribution methods apart from the interpretation of models, this paper proposes an evaluation framework for attribution methods in NLP that uses only models that are fully understood \textit{a priori}. Instead of testing attribution methods on black-box models obtained through training, we construct \textit{white-box} models for testing by directly setting network parameters by hand. Our focus is on white-box LSTMs that implement intuitive strategies for solving simple classification tasks based on formal languages with deterministic solutions. We apply our framework to five attribution methods: \textit{occlusion} \citep{zeilerVisualizingUnderstandingConvolutional2014}, \textit{saliency} \citep{simonyanDeepConvolutionalNetworks2014,liVisualizingUnderstandingNeural2016}, \textit{gradient $\times$ input}, (G $\times$ I, \citealp{shrikumarNotJustBlack2017}), \textit{integrated gradients} (IG, \citealp{sundararajanAxiomaticAttributionDeep2017}), and \textit{layer-wise relevance propagation} (LRP, \citealp{bachPixelWiseExplanationsNonLinear2015}). In doing so, we make the following contributions.
\begin{itemize}
	\item We construct four white-box LSTMs that can be used to test attribution methods. We provide a complete description of our model weights in \autoref{sec:appendixa}.\footnote{We also provide code for our models at \url{https://github.com/yidinghao/whitebox-lstm}.} Beyond the five methods considered here, our white-box networks can be used to test any attribution method compatible with LSTMs.
	
	\item Empirically, we show that all five attribution methods produce erroneous heatmaps for our white-box networks, despite the models' transparent behavior. As a preview of our results, \autoref{fig:preview} shows sample heatmaps computed for two models designed to identify the non-contiguous subsequence \ta\tb\ in the input \ttt{aacb}. Even though both models' outputs are determined by the presence of the two \ta s and the \tb, all four methods either incorrectly highlight the \tc\ or fail to highlight at least one of the \ta s  in at least one case.
	
	\item We identify two general ways in which four of the five methods do not behave as intended. Firstly, while saliency, G $\times$ I and IG are theoretically invariant to differences in model implementation \citep{sundararajanAxiomaticAttributionDeep2017}, in practice we find that these methods can still produce qualitatively different heatmaps for nearly identical models. Secondly, we find that LRP is susceptible to numerical issues, which cause heatmaps to be zeroed out when values are rounded to zero.
\end{itemize}

\begin{table}[t]
	\small
	\textbf{Task:} Determine whether the input contains one of the following subsequences: \ta\tb, \tb\tc, \tc\td, or \td\tc. \\
	\textbf{Output:} \textit{True}, since the input \ttt{aacb} contains two (non-contiguous) instances of \ta\tb.
	\begin{center}
		\begin{tabular}{ c c c c c}
			\hline
		 	\textbf{Occlusion} & \textbf{Saliency} & \textbf{G $\times$ I} & \textbf{IG} & \textbf{LRP} \\ \hline
			\textcolor[rgb]{0.8714925112588235,0.8623093793176471,0.8570162640588236}{\textbf{a}} \textcolor[rgb]{0.8714925112588235,0.8623093793176471,0.8570162640588236}{\textbf{a}} \textcolor[rgb]{0.8674276350862745,0.864376599772549,0.8626024620196079}{\textbf{c}} \textcolor[rgb]{0.705673158,0.01555616,0.150232812}{\textbf{b}}  & \textcolor[rgb]{0.8674276350862745,0.864376599772549,0.8626024620196079}{\textbf{a}} \textcolor[rgb]{0.9593847296274509,0.6103057604117648,0.4893818509411765}{\textbf{a}} \textcolor[rgb]{0.9582793979254902,0.604335096882353,0.48329710138823534}{\textbf{c}} \textcolor[rgb]{0.705673158,0.01555616,0.150232812}{\textbf{b}} & \textcolor[rgb]{0.8674276350862745,0.864376599772549,0.8626024620196079}{\textbf{a}} \textcolor[rgb]{0.8674276350862745,0.864376599772549,0.8626024620196079}{\textbf{a}} \textcolor[rgb]{0.8674276350862745,0.864376599772549,0.8626024620196079}{\textbf{c}} \textcolor[rgb]{0.705673158,0.01555616,0.150232812}{\textbf{b}}  & \textcolor[rgb]{0.9604900613294117,0.6162764239411764,0.49546660049411767}{\textbf{a}} \textcolor[rgb]{0.9604900613294117,0.6162764239411764,0.49546660049411767}{\textbf{a}} \textcolor[rgb]{0.8674276350862745,0.864376599772549,0.8626024620196079}{\textbf{c}} \textcolor[rgb]{0.705673158,0.01555616,0.150232812}{\textbf{b}}  & \textcolor[rgb]{0.8674276350862745,0.864376599772549,0.8626024620196079}{\textbf{a}} \textcolor[rgb]{0.8674276350862745,0.864376599772549,0.8626024620196079}{\textbf{a}} \textcolor[rgb]{0.8674276350862745,0.864376599772549,0.8626024620196079}{\textbf{c}} \textcolor[rgb]{0.705673158,0.01555616,0.150232812}{\textbf{b}}  \\
			\textcolor[rgb]{0.8674276350862745,0.864376599772549,0.8626024620196079}{\textbf{a}} \textcolor[rgb]{0.705673158,0.01555616,0.150232812}{\textbf{a}} \textcolor[rgb]{0.705673158,0.01555616,0.150232812}{\textbf{c}} \textcolor[rgb]{0.705673158,0.01555616,0.150232812}{\textbf{b}}  & \textcolor[rgb]{0.8674276350862745,0.864376599772549,0.8626024620196079}{\textbf{a}} \textcolor[rgb]{0.8674276350862745,0.864376599772549,0.8626024620196079}{\textbf{a}} \textcolor[rgb]{0.8674276350862745,0.864376599772549,0.8626024620196079}{\textbf{c}} \textcolor[rgb]{0.705673158,0.01555616,0.150232812}{\textbf{b}}  & \textcolor[rgb]{0.8674276350862745,0.864376599772549,0.8626024620196079}{\textbf{a}} \textcolor[rgb]{0.8674276350862745,0.864376599772549,0.8626024620196079}{\textbf{a}} \textcolor[rgb]{0.8674276350862745,0.864376599772549,0.8626024620196079}{\textbf{c}} \textcolor[rgb]{0.705673158,0.01555616,0.150232812}{\textbf{b}}  & \textcolor[rgb]{0.8674276350862745,0.864376599772549,0.8626024620196079}{\textbf{a}} \textcolor[rgb]{0.705673158,0.01555616,0.150232812}{\textbf{a}} \textcolor[rgb]{0.9695216017294117,0.7008328676235294,0.5875080175294117}{\textbf{c}} \textcolor[rgb]{0.9491505082901961,0.7907852690666667,0.7108755943019608}{\textbf{b}}  & \textcolor[rgb]{0.8674276350862745,0.864376599772549,0.8626024620196079}{\textbf{a}} \textcolor[rgb]{0.8674276350862745,0.864376599772549,0.8626024620196079}{\textbf{a}} \textcolor[rgb]{0.8674276350862745,0.864376599772549,0.8626024620196079}{\textbf{c}} \textcolor[rgb]{0.705673158,0.01555616,0.150232812}{\textbf{b}}  \\\hline
		\end{tabular}
	\end{center}
	\small
	
	\caption{Sample heatmaps for two white-box networks: a ``counter-based'' network (top) and an ``FSA-based'' network (bottom). The features relevant to the output are the two \ta s and the \tb.}
	\label{fig:preview}
\end{table}

\section{Related Work}
\label{sec:relatedwork}

Several approaches have been taken in the literature for understanding how to evaluate attribution methods. On a theoretical level, \textit{axiomatic} approaches propose formal desiderata that attribution methods should satisfy, such as implementation invariance \citep{sundararajanAxiomaticAttributionDeep2017}, input translation invariance \citep{kindermansReliabilitySaliencyMethods2019}, continuity with respect to inputs \citep{montavonMethodsInterpretingUnderstanding2018,ghorbaniInterpretationNeuralNetworks2019}, or the existence of relationships between attribution scores and logit or softmax scores \citep{sundararajanAxiomaticAttributionDeep2017,anconaBetterUnderstandingGradientbased2018,montavonGradientBasedVsPropagationBased2019}. The degree to which attribution methods fulfill these criteria can be determined either mathematically or empirically.

Other approaches, which are more experimental in nature, attempt to directly assess the relationship between attribution scores and model behavior. A common test, due to \citet{bachPixelWiseExplanationsNonLinear2015} and \citet{samekEvaluatingVisualizationWhat2017} and applied to sequence modeling by \citet{arrasWhatRelevantText2017}, involves ablating or perturbing parts of the input, from those with the highest attribution scores to those with the lowest, and counting the number of features that need to be ablated in order to change the model's prediction. Another test, proposed by \citet{adebayoSanityChecksSaliency2018}, tracks how heatmaps change as layers of a network are incrementally randomized.

A third kind of approach evaluates the extent to which heatmaps identify salient input features. For example, \citet{zhangTopDownNeuralAttention2018} propose the \textit{pointing game task}, in which the highest-relevance pixel for an image classifier input must belong to the object described by the target output class. Within this framework, \citet{kimInterpretabilityFeatureAttribution2018}, \citet{poernerEvaluatingNeuralNetwork2018}, \citet{arrasEvaluatingRecurrentNeural2019}, and \citet{yangBenchmarkingAttributionMethods2019} construct datasets in which input features exhibit experimentally controlled notions of importance, yielding ``ground truth'' attributions against which heatmaps can be evaluated. 

Our paper incorporates elements of the ground-truth approaches, since it is straightforward to determine which input features are important for our formal language tasks. We enhance these approaches by using white-box models that are guaranteed to be sensitive to those features.

\section{Formal Language Tasks}
\label{sec:tasks}

Formal languages are often used to evaluate the expressive power of RNNs. Here, we focus on formal languages that have been recently used to probe LSTMs' ability to capture three kinds of dependencies: \textit{counting}, \textit{long-distance}, and \textit{hierarchical} dependencies. We define a classification task based on each of these formal languages.

\subsection{Counting Dependencies}

\textit{Counter languages} \citep{fischerTuringMachinesRestricted1966,fischerCounterMachinesCounter1968} are languages recognized by automata equipped with counters. \citet{weissPracticalComputationalPower2018a} demonstrate using an acceptance task for the languages $\ta^n\tb^n$ and $\ta^n\tb^n\tc^n$ that LSTMs naturally learn to use cell state units as counters. \citeauthor{merrillSequentialNeuralNetworks2019a}'s (\citeyear{merrillSequentialNeuralNetworks2019a}) asymptotic analysis shows that LSTM acceptors accept only counter languages when their weights are fully saturated. Thus, counter languages may be viewed as a characterization of the expressive power of LSTMs. 

We define the \textit{counting task} based on a simple example of a counting language.
\begin{task}[Counting Task]
	Given a string in $x \in \lbrace \ta, \tb \rbrace^*$, determine whether or not $x$ has strictly more \ta s than \tb s.
\end{task}
\begin{example}
	The counting task classifies \ttt{aaab} as \textit{True}, \ttt{ab} as \textit{False}, and \ttt{bbbba} as \textit{False}.
\end{example}
A counter automaton can solve the counting task by incrementing its counter whenever an \ta\ is encountered and decrementing it whenever a \tb\ is encountered. It outputs \textit{True} if and only if its counter is at least $1$. We expect attribution scores for all input symbols to have roughly the same magnitude, but that scores assigned to \ta\ will have the opposite sign to those assigned to \tb.

\subsection{Long-Distance Dependencies}

\textit{Strictly piecewise} (SP, \citealp{heinzInductiveLearningPhonotactic2007}) languages were used by \citet{avcuSubregularComplexityDeep2017} and \citet{mahalunkarUsingRegularLanguages2018,mahalunkarMultiElementLongDistance2019,mahalunkarUnderstandingRecurrentNeural2019} to test the propensity of LSTMs to learn long-distance dependencies, compared to \citeauthor{elmanFindingStructureTime1990}'s (\citeyear{elmanFindingStructureTime1990}) simple recurrent networks. SP languages are regular languages whose membership is defined by the presence or absence of certain \textit{subsequences}, which may or may not be contiguous. For example, \ttt{ad} is a subsequence of \ttt{abcde}, since both letters of \ttt{ad} occur in \ttt{abcde}, in the same order. Based on these ideas, we define the \textit{SP task} as follows.
\begin{task}[SP Task]
	Given $x \in \lbrace \ta, \tb, \tc, \td \rbrace^*$, determine whether or not $x$ contains at least one of the following subsequences: \ttt{ab}, \ttt{bc}, \ttt{cd}, \ttt{dc}.
\end{task}
\begin{example}
	In the SP task, \ttt{aab} is classified as \textit{True}, since it contains the subsequence \ttt{ab}. Similarly, \ttt{acb} is classified as \textit{True}, since it contains \ttt{ab} non-contiguously. The string \ttt{aaa} is classified as \textit{False}.
\end{example}

The choice of SP languages as a test for long-distance dependencies is motivated by the fact that symbols in a non-contiguous subsequence may occur arbitrarily far from one another. The SP task yields a variant of the pointing game task in the sense that the input string may or may not contain an ``object'' (one of the four subsequences) that the network must identify. Therefore, we expect an input symbol to receive a nonzero attribution score if and only if it comprises a subsequence.

\subsection{Hierarchical Dependencies}

The \textit{Dyck language} is the language $D$ generated by the following context-free grammar, where $\varepsilon$ is the empty string.
\[
S \to SS \mathrel{|} \lp S \rp \mathrel{|} \lb S \rb \mathrel{|} \varepsilon
\]
$D$ contains all balanced strings of parentheses and square brackets. Since $D$ is often viewed as a canonical example of a context-free language \citep{chomskyAlgebraicTheoryContextFree1959}, several recent studies, including \citet{sennhauserEvaluatingAbilityLSTMs2018}, \citet{bernardyCanRecurrentNeural2018}, \citet{skachkovaClosingBracketsRecurrent2018}, and \citet{yuLearningDyckLanguage2019}, have used $D$ to evaluate whether LSTMs can learn hierarchical dependencies implemented by pushdown automata. Here, we consider the \textit{bracket prediction task} proposed by \citet{sennhauserEvaluatingAbilityLSTMs2018}. 

\begin{task}[Bracket Prediction Task]
	Given a prefix $p$ of some string in $D$, identify the next valid closing bracket for $p$. 
\end{task}
\begin{example}
	The string \texttt{[([]} requires a prediction of \texttt{)}, since the \texttt{(} is the last unclosed bracket. Similarly, \texttt{(()[} requires a prediction of \texttt{]}. Strings with no unclosed brackets, such as \texttt{[()]}, require a prediction of \textit{None}.
\end{example}

In heatmaps for the bracket prediction task, we expect the last unclosed bracket to receive the highest-magnitude relevance score.

\section{White-Box Networks}
\label{sec:whiteboxnetworks}

We use two approaches to construct white-box networks for our tasks. In the \textit{counter-based} approach, the cell state contains a set of counters, which are incremented or decremented throughout the computation. The network's final output is based on the values of the counters. In the \textit{automaton-based} approach, we use the LSTM to simulate an automaton, with the cell state containing a representation of the automaton's state. We use a counter-based network to solve the counter task and an automaton-based network to solve the bracket prediction task. We use both kinds of networks to solve the SP task. All networks perfectly solve the tasks they were designed for. This section describes our white-box networks at a high level; a detailed description is given in \autoref{sec:appendixa}.

In the rest of this paper, we identify the alphabet symbols \ta, \tb, \tc, and \td\ with the one-hot vectors for indices $1$, $2$, $3$, and $4$, respectively. The vectors $\bm{f}^{(t)}$, $\bm{i}^{(t)}$, and $\bm{o}^{(t)}$ represent the forget, input, and output gates, respectively. $\bm{g}^{(t)}$ is the value added to the cell state at each time step, and $\sigma$ represents the sigmoid function. We assume that the hidden state $\bm{h}^{(t)}$ and cell state $\bm{c}^{(t)}$ are updated as follows.
\begin{align*}
	\bm{c}^{(t)} &= \bm{f}^{(t)} \odot \bm{c}^{(t - 1)} + \bm{i}^{(t)} \odot \bm{g}^{(t)} \\
	\bm{h}^{(t)} &= \bm{o}^{(t)} \odot \tanh\left(\bm{c}^{(t)}\right)
\end{align*}

\subsection{Counter-Based Networks}
\label{sec:counternetworks}

In the counter-based approach, each position of the cell state contains the value of a counter. To adjust the counter in position $j$ by some value $v \in (-1, 1)$, we set $g_j^{(t)} = v$, and we saturate the gates by setting them to $\sigma(m) \approx 1$, where $m \gg 0$ is a large constant. For example, our network for the counting task uses a single hidden unit, with the gates always saturated and with $g^{(t)}$ given by
\[
g^{(t)} = \tanh\left(u \left[
\begin{array}{c c}
1 & -1
\end{array}
\right] \bm{x}^{(t)} \right)\text{,}
\]
where $u > 0$ is a hyperparameter that scales the counter by a factor of $v = \tanh(u)$.\footnote{We use $u = 0.5$ for the counting task, $u = 0.7$ for the SP task, and $m = 50$ for both tasks.} When $\bm{x}^{(t)} = \ta$, we have $g^{(t)} = v$, so the counter is incremented by $v$. When $\bm{x}^{(t)} = \tb$, we compute $g^{(t)} = -v$, so the counter is decremented by $v$.

For the SP task, we use seven counters. The first four counters record how many occurrences of each symbol have been observed at time step $t$. The next three counters record the number of \tb s, \tc s, and \td s that form one of the four distinguished subsequences with an earlier symbol. For example, after seeing the input \ttt{aaabbc}, the counter-based network for the SP task satisfies
\[
\bm{c}^{(6)} = v \left[\begin{array}{ccccccc}
3 & 2 & 1 & 0 & 2 & 1 & 0
\end{array}\right]^\top\text{.}
\]
The first four counters represent the fact that the input has 3 \ta s, 2 \tb s, 1 \tc, and no \td s. Counter \#5 is $2v$ because the two \tb s form a subsequence with the \ta s, and counter \#6 is $v$ because the \tc\ forms a subsequence with the \tb s.

The logit scores of our counter-based networks are computed by a linear decoder using the $\tanh$ of the counter values. For the counting task, the score of the \textit{True} class is $h^{(t)}$, while the score of the \textit{False} class is fixed to $\tanh(v)/2$. This means that the network outputs \textit{True} if and only if the final counter value is at least $v$. For the SP task, the score of the \textit{True} class is $h_5^{(t)} + h_6^{(t)} + h_7^{(t)}$, while the score of the \textit{False} class is again $\tanh(v)/2$.

\subsection{Automata-Based Networks}
\label{sec:automatanetworks}

We consider two types of automata-based networks: one that implements a finite-state automaton (FSA) for the SP task, and one that implements a pushdown automaton (PDA) for the bracket prediction task.

Our FSA construction is similar to \citeauthor{korskyComputationalPowerRNNs2019}'s (\citeyear{korskyComputationalPowerRNNs2019}) FSA construction for simple recurrent networks. Consider a deterministic FSA $\mathcal{A}$ with states $Q$ and alphabet $\Sigma$. To simulate $\mathcal{A}$ using an LSTM, we use $|Q| \cdot |\Sigma|$ hidden units, with the following interpretation. Suppose that $\mathcal{A}$ transitions to state $q$ after reading input $\bm{x}^{(1)}, \bm{x}^{(2)}, \dots, \bm{x}^{(t)}$. The hidden state $\bm{h}^{(t)}$ is a one-hot representation of the pair $\left\langle q, \bm{x}^{(t)} \right\rangle$, which encodes both the current state of $\mathcal{A}$ and the most recent input symbol. Since the FSA undergoes a state transition with each input symbol, the forget gate always clears $\bm{c}^{(t)}$, so that information written to the cell state does not persist beyond a single time step. The output layer simply detects whether or not the FSA is in an accepting state. Details are provided in \autoref{sec:fsadetails}.

Next, we describe how to implement a PDA for the bracket prediction task. We use a stack containing all unclosed brackets observed in the input string, and make predictions based on the top item of the stack. We represent a bounded stack of size $k$ using $2k + 1$ hidden units. The first $k - 1$ positions contain all stack items except the top item, with \texttt{(} represented by the value $1$, \texttt{[} represented by $-1$, and empty positions represented by $0$. The $k$th position contains the top item of the stack. The next $k$ positions contain the height of the stack in unary notation, and the last position contains a bit indicating whether or not the stack is empty. For example, after reading the input \texttt{([(()} with a stack of size 4, the stack contents \texttt{([(} are represented by
\[
\bm{c}^{(5)} = \left[ \begin{array}{c c c c c c c c c}
1 & -1 & 0 & 1 & 1 & 1 & 1 & 0 & 0
\end{array}\right]^\top\text{.}
\]
The $1$ in position 4 indicates that the top item of the stack is \texttt{(}, and the $1$, $-1$, and $0$ in positions 1--3 indicate that the remainder of the stack is \texttt{([}. The three $1$s in positions 5--8 indicate that the stack height is 3, and the 0 in position 9 indicates that the stack is not empty.

When $\bm{x}^{(t)}$ is \texttt{(} or \texttt{[}, it is copied to $c_k^{(t)}$, and $c_k^{(t)}$ is copied to the highest empty position in $\bm{c}_{:k - 1}^{(t)}$, pushing the opening bracket to the stack. The empty stack bit is then set to $0$, marking the stack as non-empty. When the current input symbol is a closing bracket, the highest item of positions 1 through $k - 1$ is deleted and copied to position $k$, popping the top item from the stack. Because the PDA network is quite complex, we focus here on describing how the top stack item in position $k$ is determined, and leave other details for \autoref{sec:pdaappendix}. Let $\alpha^{(t)}$ be $1$ if $\bm{x}^{(t)} = \texttt{(}$, $-1$ if $\bm{x}^{(t)} = \texttt{[}$, and $0$ otherwise. At each time step, $g^{(t)}_k = \tanh\left(m \cdot u^{(t)}\right)$, where $m \gg 0$ and 
\begin{equation}
u^{(t)} = 2^k\alpha^{(t)} + \sum_{j = 1}^{k - 1} 2^{j - 1}h_j^{(t - 1)}\text{.} \label{eqn:bracketnetwork}
\end{equation}
Observe that $m \cdot u^{(t)} \gg 0$ when $\alpha^{(t)} = 1$, and $m \cdot u^{(t)} \ll 0$ when $\alpha^{(t)} = -1$. Thus, $g_k^{(t)}$ contains the stack encoding of the current input symbol if it is an opening bracket. If the current input symbol is a closing bracket, then  $\alpha^{(t)} = 0$, so the sign of $u^{(t)}$ is determined by the highest item of $\bm{h}_{:k - 1}^{(t - 1)}$.

\section{Attribution Methods}	
\label{sec:attributionmethods}

\begin{table}
	\centering
	\small
	\begin{tabular}{l l}
		\hline
		\textbf{Name} & \textbf{Formula} \\\hline \vspace{-.75em} \\
		Saliency & $\displaystyle R^{(c)}_{t, i}(\bm{X}) = \frac{\partial \hat{y}_c}{\partial x^{(t)}_i}\Bigr|_{x^{(t)}_i = X_{t, i}}$ \\ \vspace{-.5em} \\
		G $\times$ I & $\displaystyle R^{(c)}_{t, i}(\bm{X}) = X_{t, i} \frac{\partial \hat{y}_c}{\partial x^{(t)}_i}\Bigr|_{x^{(t)}_i = X_{t, i}}$ \\ \vspace{-.5em} \\
		IG & $\displaystyle R^{(c)}_{t, i}(\bm{X}) = X_{t, i} \int_0^1 \frac{\partial \hat{y}_c}{\partial x^{(t)}_i}\Bigr|_{x^{(t)}_i = \alpha X_{t, i}} \, d\alpha$ \\ \vspace{-.75em} \\\hline 
	\end{tabular}
	\caption{Definitions of the gradient-based methods.}
	\label{fig:attributionmethods}
\end{table}

Let $\bm{X}$ be a matrix of input vectors, such that the input at time $t$ is the row vector $\bm{X}_{t, :} = \left(\bm{x}^{(t)}\right)^\top$. Given $\bm{X}$, an LSTM classifier produces a vector $\hat{\bm{y}}$ of logit scores. Based on $\bm{X}$, $\hat{\bm{y}}$, and possibly a \textit{baseline input} $\overline{\bm{X}}$, an attribution method assigns an attribution score $R^{(c)}_{t, i}(\bm{X})$ to input feature $X_{t, i}$ for each output class $c$. These feature-level scores are then aggregated to produce token-level scores:
\[
R^{(c)}_t(\bm{X}) = \sum_{i} R^{(c)}_{t, i}(\bm{X})\text{.}
\]
Broadly speaking, our five attribution methods are grouped into three types: one \textit{perturbation-based}, three \textit{gradient-based}, and one \textit{decomposition-based}. The following subsections describe how each method computes $R^{(c)}_{t, i}(\bm{X})$.

\subsection{Perturbation- and Gradient-Based Methods}

Perturbation-based methods are premised on the idea that if $X_{t, i}$ is an important input feature, then changing the value of $X_{t, i}$ would cause $\hat{\bm{y}}$ to change. The one perturbation method we consider is occlusion. In this method, $R^{(c)}_{t, i}(\bm{X})$ is the change in $\hat{y}_c$ observed when $\bm{X}_{t, :}$ is replaced by $\bm{0}$. 

Gradient-based methods rely on the same intuition as perturbation-based methods, but use automatic differentiation to simulate infinitesimal perturbations. The definitions of our three gradient-based methods are given in \autoref{fig:attributionmethods}. The most basic of these is saliency, which simply measures relevance by the derivative of the logit score with respect to each input feature. G $\times$ I attempts to improve upon saliency by using the first-order terms in a Taylor-series approximation of the model instead of the gradients on their own. IG is designed to address the issue of small gradients found in saturated units by integrating G $\times$ I along the line connecting $\bm{X}$ to a baseline input $\overline{\bm{X}}$, here taken to be the zero matrix.

\subsection{Decomposition-Based Methods}

Decomposition-based methods are methods that satisfy the relation
\begin{equation}
\hat{y}_c = R^{(c)}_{\text{bias}} + \sum_{t, i} R^{(c)}_{t, i}(\bm{X})\text{,} \label{eqn:conservation}
\end{equation}
where $R^{(c)}_{\text{bias}}$ is a relevance score assigned to the bias units of the network. The interpretation of \autoref{eqn:conservation} is that the logit score $\hat{y}_c$ is ``distributed'' among the input features and the bias units, so that the relevance scores form a ``decomposition'' of $\hat{y}_c$.

The one decomposition-based method we consider is LRP, which computes scores using a backpropagation algorithm that distributes scores layer by layer. The scores of the output layer are initialized to
\[
r^{(c, \text{output})}_i = \begin{cases}
\hat{y}_i, & i = c \\
0, & \text{otherwise.}
\end{cases}
\]
For each layer $l$ with activation $\bm{z}^{(l)}$, activation function $f^{(l)}$, and output $\bm{a}^{(l)} = f^{(l)}\left(\bm{z}^{(l)}\right)$, the relevance $\bm{r}^{(c, l)}$ of $\bm{a}^{(l)}$ is determined by the following \textit{propagation rule}:
\[
r^{(c, l)}_i = \sum_{l^\prime} \sum_j r^{(c, l^\prime)}_j \frac{W^{(l^\prime \gets l)}_{j, i}a^{(l)}_i}{z^{(l^\prime)}_j + \sign\left(z^{(l^\prime)}_j\right) \varepsilon }\text{,}
\]
where $l^\prime$ ranges over all layers to which $l$ has a forward connection via $\bm{W}^{(l^\prime \gets l)}$ and $\varepsilon > 0$ is a stabilizing constant.\footnote{We use $\varepsilon = 0.001$.} For the LSTM gate interactions, we follow \citet{arrasExplainingRecurrentNeural2017} in treating multiplicative connections of the form $\bm{a}^{(l_1)} \odot \bm{a}^{(l_2)}$ as activation functions of the form $\bm{a}^{(l_1)} \odot f^{(l_2)}(\cdot)$, where $\bm{a}^{(l_1)}$ is $\bm{f}^{(t)}$, $\bm{i}^{(t)}$, or $\bm{o}^{(t)}$. The final attribution scores are given by the values propagated to the input layer: 
\[
R^{(c)}_{t, i}(\bm{X}) = r_i^{(c, \text{input}_t)}\text{.}
\]

%

\section{Qualitative Evaluation}
\label{sec:qualitativeevaluation}

\newcounter{heatmapnum}
\newcommand{\heatmaprow}[2]{\stepcounter{heatmapnum}&\theheatmapnum&#1&#2&}
\begin{table*}[h]
	\small
	\centering
	\begin{tabular}{l c c c l l l l l}
		\hline
		\textbf{Network} & \textbf{\#} & $c$ & \textbf{Target} & \textbf{Occlusion} & \textbf{Saliency} & \textbf{G $\times$ I} & \textbf{IG} & \textbf{LRP} \\\hline
		
		\multirow{6}{*}{Counting}
		\heatmaprow{True}{True} \textcolor[rgb]{0.705673158,0.01555616,0.150232812}{\textbf{a}} \textcolor[rgb]{0.705673158,0.01555616,0.150232812}{\textbf{a}} \textcolor[rgb]{0.705673158,0.01555616,0.150232812}{\textbf{a}} \textcolor[rgb]{0.42519897019607844,0.559058179764706,0.9460614570784314}{\textbf{b}} \textcolor[rgb]{0.42519897019607844,0.559058179764706,0.9460614570784314}{\textbf{b}}  & \textcolor[rgb]{0.8674276350862745,0.864376599772549,0.8626024620196079}{\textbf{a}} \textcolor[rgb]{0.8674276350862745,0.864376599772549,0.8626024620196079}{\textbf{a}} \textcolor[rgb]{0.8674276350862745,0.864376599772549,0.8626024620196079}{\textbf{a}} \textcolor[rgb]{0.8674276350862745,0.864376599772549,0.8626024620196079}{\textbf{b}} \textcolor[rgb]{0.8674276350862745,0.864376599772549,0.8626024620196079}{\textbf{b}}  & \textcolor[rgb]{0.705673158,0.01555616,0.150232812}{\textbf{a}} \textcolor[rgb]{0.705673158,0.01555616,0.150232812}{\textbf{a}} \textcolor[rgb]{0.705673158,0.01555616,0.150232812}{\textbf{a}} \textcolor[rgb]{0.2298057,0.298717966,0.753683153}{\textbf{b}} \textcolor[rgb]{0.2298057,0.298717966,0.753683153}{\textbf{b}}  & \textcolor[rgb]{0.705673158,0.01555616,0.150232812}{\textbf{a}} \textcolor[rgb]{0.705673158,0.01555616,0.150232812}{\textbf{a}} \textcolor[rgb]{0.705673158,0.01555616,0.150232812}{\textbf{a}} \textcolor[rgb]{0.2298057,0.298717966,0.753683153}{\textbf{b}} \textcolor[rgb]{0.2298057,0.298717966,0.753683153}{\textbf{b}}  & \textcolor[rgb]{0.705673158,0.01555616,0.150232812}{\textbf{a}} \textcolor[rgb]{0.705673158,0.01555616,0.150232812}{\textbf{a}} \textcolor[rgb]{0.705673158,0.01555616,0.150232812}{\textbf{a}} \textcolor[rgb]{0.2298057,0.298717966,0.753683153}{\textbf{b}} \textcolor[rgb]{0.2298057,0.298717966,0.753683153}{\textbf{b}}  \\
		\heatmaprow{True}{False} \textcolor[rgb]{0.2298057,0.298717966,0.753683153}{\textbf{b}} \textcolor[rgb]{0.2298057,0.298717966,0.753683153}{\textbf{b}} \textcolor[rgb]{0.2298057,0.298717966,0.753683153}{\textbf{b}} \textcolor[rgb]{0.8995343807254902,0.4406918021568627,0.34410686323529416}{\textbf{a}} \textcolor[rgb]{0.8995343807254902,0.4406918021568627,0.34410686323529416}{\textbf{a}}  & \textcolor[rgb]{0.8674276350862745,0.864376599772549,0.8626024620196079}{\textbf{b}} \textcolor[rgb]{0.8674276350862745,0.864376599772549,0.8626024620196079}{\textbf{b}} \textcolor[rgb]{0.8674276350862745,0.864376599772549,0.8626024620196079}{\textbf{b}} \textcolor[rgb]{0.8674276350862745,0.864376599772549,0.8626024620196079}{\textbf{a}} \textcolor[rgb]{0.8674276350862745,0.864376599772549,0.8626024620196079}{\textbf{a}}   & \textcolor[rgb]{0.2298057,0.298717966,0.753683153}{\textbf{b}} \textcolor[rgb]{0.2298057,0.298717966,0.753683153}{\textbf{b}} \textcolor[rgb]{0.2298057,0.298717966,0.753683153}{\textbf{b}} \textcolor[rgb]{0.705673158,0.01555616,0.150232812}{\textbf{a}} \textcolor[rgb]{0.705673158,0.01555616,0.150232812}{\textbf{a}}  & \textcolor[rgb]{0.2298057,0.298717966,0.753683153}{\textbf{b}} \textcolor[rgb]{0.2298057,0.298717966,0.753683153}{\textbf{b}} \textcolor[rgb]{0.2298057,0.298717966,0.753683153}{\textbf{b}} \textcolor[rgb]{0.705673158,0.01555616,0.150232812}{\textbf{a}} \textcolor[rgb]{0.705673158,0.01555616,0.150232812}{\textbf{a}}  & \textcolor[rgb]{0.2298057,0.298717966,0.753683153}{\textbf{b}} \textcolor[rgb]{0.2298057,0.298717966,0.753683153}{\textbf{b}} \textcolor[rgb]{0.2298057,0.298717966,0.753683153}{\textbf{b}} \textcolor[rgb]{0.705673158,0.01555616,0.150232812}{\textbf{a}} \textcolor[rgb]{0.705673158,0.01555616,0.150232812}{\textbf{a}}  \\
		\heatmaprow{True}{False} \textcolor[rgb]{0.705673158,0.01555616,0.150232812}{\textbf{a}} \textcolor[rgb]{0.705673158,0.01555616,0.150232812}{\textbf{a}} \textcolor[rgb]{0.705673158,0.01555616,0.150232812}{\textbf{a}} \textcolor[rgb]{0.2298057,0.298717966,0.753683153}{\textbf{b}} \textcolor[rgb]{0.2298057,0.298717966,0.753683153}{\textbf{b}} \textcolor[rgb]{0.2298057,0.298717966,0.753683153}{\textbf{b}}  & \textcolor[rgb]{0.8674276350862745,0.864376599772549,0.8626024620196079}{\textbf{a}} \textcolor[rgb]{0.8674276350862745,0.864376599772549,0.8626024620196079}{\textbf{a}} \textcolor[rgb]{0.8674276350862745,0.864376599772549,0.8626024620196079}{\textbf{a}} \textcolor[rgb]{0.8674276350862745,0.864376599772549,0.8626024620196079}{\textbf{b}} \textcolor[rgb]{0.8674276350862745,0.864376599772549,0.8626024620196079}{\textbf{b}} \textcolor[rgb]{0.8674276350862745,0.864376599772549,0.8626024620196079}{\textbf{b}}  & \textcolor[rgb]{0.705673158,0.01555616,0.150232812}{\textbf{a}} \textcolor[rgb]{0.705673158,0.01555616,0.150232812}{\textbf{a}} \textcolor[rgb]{0.705673158,0.01555616,0.150232812}{\textbf{a}} \textcolor[rgb]{0.2298057,0.298717966,0.753683153}{\textbf{b}} \textcolor[rgb]{0.2298057,0.298717966,0.753683153}{\textbf{b}} \textcolor[rgb]{0.2298057,0.298717966,0.753683153}{\textbf{b}}  & \textcolor[rgb]{0.705673158,0.01555616,0.150232812}{\textbf{a}} \textcolor[rgb]{0.705673158,0.01555616,0.150232812}{\textbf{a}} \textcolor[rgb]{0.705673158,0.01555616,0.150232812}{\textbf{a}} \textcolor[rgb]{0.2298057,0.298717966,0.753683153}{\textbf{b}} \textcolor[rgb]{0.2298057,0.298717966,0.753683153}{\textbf{b}} \textcolor[rgb]{0.2298057,0.298717966,0.753683153}{\textbf{b}}  & \textcolor[rgb]{0.8674276350862745,0.864376599772549,0.8626024620196079}{\textbf{a}} \textcolor[rgb]{0.8674276350862745,0.864376599772549,0.8626024620196079}{\textbf{a}} \textcolor[rgb]{0.8674276350862745,0.864376599772549,0.8626024620196079}{\textbf{a}} \textcolor[rgb]{0.8674276350862745,0.864376599772549,0.8626024620196079}{\textbf{b}} \textcolor[rgb]{0.8674276350862745,0.864376599772549,0.8626024620196079}{\textbf{b}} \textcolor[rgb]{0.8674276350862745,0.864376599772549,0.8626024620196079}{\textbf{b}}  \\
		\heatmaprow{True}{False} \textcolor[rgb]{0.8995343807254902,0.4406918021568627,0.34410686323529416}{\textbf{a}} \textcolor[rgb]{0.8995343807254902,0.4406918021568627,0.34410686323529416}{\textbf{a}} \textcolor[rgb]{0.2298057,0.298717966,0.753683153}{\textbf{b}} \textcolor[rgb]{0.2298057,0.298717966,0.753683153}{\textbf{b}} \textcolor[rgb]{0.2298057,0.298717966,0.753683153}{\textbf{b}}  & \textcolor[rgb]{0.8674276350862745,0.864376599772549,0.8626024620196079}{\textbf{a}} \textcolor[rgb]{0.8674276350862745,0.864376599772549,0.8626024620196079}{\textbf{a}} \textcolor[rgb]{0.8674276350862745,0.864376599772549,0.8626024620196079}{\textbf{b}} \textcolor[rgb]{0.8674276350862745,0.864376599772549,0.8626024620196079}{\textbf{b}} \textcolor[rgb]{0.8674276350862745,0.864376599772549,0.8626024620196079}{\textbf{b}} & \textcolor[rgb]{0.705673158,0.01555616,0.150232812}{\textbf{a}} \textcolor[rgb]{0.705673158,0.01555616,0.150232812}{\textbf{a}} \textcolor[rgb]{0.2298057,0.298717966,0.753683153}{\textbf{b}} \textcolor[rgb]{0.2298057,0.298717966,0.753683153}{\textbf{b}} \textcolor[rgb]{0.2298057,0.298717966,0.753683153}{\textbf{b}}  & \textcolor[rgb]{0.705673158,0.01555616,0.150232812}{\textbf{a}} \textcolor[rgb]{0.705673158,0.01555616,0.150232812}{\textbf{a}} \textcolor[rgb]{0.2298057,0.298717966,0.753683153}{\textbf{b}} \textcolor[rgb]{0.2298057,0.298717966,0.753683153}{\textbf{b}} \textcolor[rgb]{0.2298057,0.298717966,0.753683153}{\textbf{b}}  & \textcolor[rgb]{0.8674276350862745,0.864376599772549,0.8626024620196079}{\textbf{a}} \textcolor[rgb]{0.8674276350862745,0.864376599772549,0.8626024620196079}{\textbf{a}} \textcolor[rgb]{0.8674276350862745,0.864376599772549,0.8626024620196079}{\textbf{b}} \textcolor[rgb]{0.8674276350862745,0.864376599772549,0.8626024620196079}{\textbf{b}} \textcolor[rgb]{0.2298057,0.298717966,0.753683153}{\textbf{b}}  \\
		\heatmaprow{False}{True} \textcolor[rgb]{0.8674276350862745,0.864376599772549,0.8626024620196079}{\textbf{a}} \textcolor[rgb]{0.8674276350862745,0.864376599772549,0.8626024620196079}{\textbf{a}} \textcolor[rgb]{0.8674276350862745,0.864376599772549,0.8626024620196079}{\textbf{a}} \textcolor[rgb]{0.8674276350862745,0.864376599772549,0.8626024620196079}{\textbf{b}} \textcolor[rgb]{0.8674276350862745,0.864376599772549,0.8626024620196079}{\textbf{b}}  & \textcolor[rgb]{0.8674276350862745,0.864376599772549,0.8626024620196079}{\textbf{a}} \textcolor[rgb]{0.8674276350862745,0.864376599772549,0.8626024620196079}{\textbf{a}} \textcolor[rgb]{0.8674276350862745,0.864376599772549,0.8626024620196079}{\textbf{a}} \textcolor[rgb]{0.8674276350862745,0.864376599772549,0.8626024620196079}{\textbf{b}} \textcolor[rgb]{0.8674276350862745,0.864376599772549,0.8626024620196079}{\textbf{b}}  & \textcolor[rgb]{0.8674276350862745,0.864376599772549,0.8626024620196079}{\textbf{a}} \textcolor[rgb]{0.8674276350862745,0.864376599772549,0.8626024620196079}{\textbf{a}} \textcolor[rgb]{0.8674276350862745,0.864376599772549,0.8626024620196079}{\textbf{a}} \textcolor[rgb]{0.8674276350862745,0.864376599772549,0.8626024620196079}{\textbf{b}} \textcolor[rgb]{0.8674276350862745,0.864376599772549,0.8626024620196079}{\textbf{b}}  & \textcolor[rgb]{0.8674276350862745,0.864376599772549,0.8626024620196079}{\textbf{a}} \textcolor[rgb]{0.8674276350862745,0.864376599772549,0.8626024620196079}{\textbf{a}} \textcolor[rgb]{0.8674276350862745,0.864376599772549,0.8626024620196079}{\textbf{a}} \textcolor[rgb]{0.8674276350862745,0.864376599772549,0.8626024620196079}{\textbf{b}} \textcolor[rgb]{0.8674276350862745,0.864376599772549,0.8626024620196079}{\textbf{b}}  & \textcolor[rgb]{0.8674276350862745,0.864376599772549,0.8626024620196079}{\textbf{a}} \textcolor[rgb]{0.8674276350862745,0.864376599772549,0.8626024620196079}{\textbf{a}} \textcolor[rgb]{0.8674276350862745,0.864376599772549,0.8626024620196079}{\textbf{a}} \textcolor[rgb]{0.8674276350862745,0.864376599772549,0.8626024620196079}{\textbf{b}} \textcolor[rgb]{0.8674276350862745,0.864376599772549,0.8626024620196079}{\textbf{b}}  \\
		\heatmaprow{False}{False} \textcolor[rgb]{0.8674276350862745,0.864376599772549,0.8626024620196079}{\textbf{a}} \textcolor[rgb]{0.8674276350862745,0.864376599772549,0.8626024620196079}{\textbf{a}} \textcolor[rgb]{0.8674276350862745,0.864376599772549,0.8626024620196079}{\textbf{b}} \textcolor[rgb]{0.8674276350862745,0.864376599772549,0.8626024620196079}{\textbf{b}} \textcolor[rgb]{0.8674276350862745,0.864376599772549,0.8626024620196079}{\textbf{b}}  & \textcolor[rgb]{0.8674276350862745,0.864376599772549,0.8626024620196079}{\textbf{a}} \textcolor[rgb]{0.8674276350862745,0.864376599772549,0.8626024620196079}{\textbf{a}} \textcolor[rgb]{0.8674276350862745,0.864376599772549,0.8626024620196079}{\textbf{b}} \textcolor[rgb]{0.8674276350862745,0.864376599772549,0.8626024620196079}{\textbf{b}} \textcolor[rgb]{0.8674276350862745,0.864376599772549,0.8626024620196079}{\textbf{b}}  & \textcolor[rgb]{0.8674276350862745,0.864376599772549,0.8626024620196079}{\textbf{a}} \textcolor[rgb]{0.8674276350862745,0.864376599772549,0.8626024620196079}{\textbf{a}} \textcolor[rgb]{0.8674276350862745,0.864376599772549,0.8626024620196079}{\textbf{b}} \textcolor[rgb]{0.8674276350862745,0.864376599772549,0.8626024620196079}{\textbf{b}} \textcolor[rgb]{0.8674276350862745,0.864376599772549,0.8626024620196079}{\textbf{b}}  & \textcolor[rgb]{0.8674276350862745,0.864376599772549,0.8626024620196079}{\textbf{a}} \textcolor[rgb]{0.8674276350862745,0.864376599772549,0.8626024620196079}{\textbf{a}} \textcolor[rgb]{0.8674276350862745,0.864376599772549,0.8626024620196079}{\textbf{b}} \textcolor[rgb]{0.8674276350862745,0.864376599772549,0.8626024620196079}{\textbf{b}} \textcolor[rgb]{0.8674276350862745,0.864376599772549,0.8626024620196079}{\textbf{b}}  & \textcolor[rgb]{0.8674276350862745,0.864376599772549,0.8626024620196079}{\textbf{a}} \textcolor[rgb]{0.8674276350862745,0.864376599772549,0.8626024620196079}{\textbf{a}} \textcolor[rgb]{0.8674276350862745,0.864376599772549,0.8626024620196079}{\textbf{b}} \textcolor[rgb]{0.8674276350862745,0.864376599772549,0.8626024620196079}{\textbf{b}} \textcolor[rgb]{0.8674276350862745,0.864376599772549,0.8626024620196079}{\textbf{b}}  \\\hline
		
		\multirow{7}{*}{SP (Counter)} 
		\heatmaprow{True}{True} \textcolor[rgb]{0.705673158,0.01555616,0.150232812}{\textbf{a}} \textcolor[rgb]{0.8674276350862745,0.864376599772549,0.8626024620196079}{\textbf{c}} \textcolor[rgb]{0.705673158,0.01555616,0.150232812}{\textbf{b}}  & \textcolor[rgb]{0.9563709270509804,0.7751443261333334,0.6864159483098039}{\textbf{a}} \textcolor[rgb]{0.8008302512666666,0.2508292108,0.22569570326666666}{\textbf{c}} \textcolor[rgb]{0.705673158,0.01555616,0.150232812}{\textbf{b}}   & \textcolor[rgb]{0.8610536002941176,0.3629157635294118,0.2906281271764706}{\textbf{a}} \textcolor[rgb]{0.8674276350862745,0.864376599772549,0.8626024620196079}{\textbf{c}} \textcolor[rgb]{0.705673158,0.01555616,0.150232812}{\textbf{b}}  & \textcolor[rgb]{0.705673158,0.01555616,0.150232812}{\textbf{a}} \textcolor[rgb]{0.8674276350862745,0.864376599772549,0.8626024620196079}{\textbf{c}} \textcolor[rgb]{0.9094595977529412,0.8393864797647058,0.8003313524235294}{\textbf{b}}  & \textcolor[rgb]{0.8674276350862745,0.864376599772549,0.8626024620196079}{\textbf{a}} \textcolor[rgb]{0.8674276350862745,0.864376599772549,0.8626024620196079}{\textbf{c}} \textcolor[rgb]{0.705673158,0.01555616,0.150232812}{\textbf{b}}  \\
		\heatmaprow{True}{True} \textcolor[rgb]{0.705673158,0.01555616,0.150232812}{\textbf{a}} \textcolor[rgb]{0.8674276350862745,0.864376599772549,0.8626024620196079}{\textbf{c}} \textcolor[rgb]{0.9691920510470589,0.705835675717647,0.5937042609803921}{\textbf{b}} \textcolor[rgb]{0.9691920510470589,0.705835675717647,0.5937042609803921}{\textbf{b}}  & \textcolor[rgb]{0.9281160096666666,0.8221971488627451,0.765141349254902}{\textbf{a}} \textcolor[rgb]{0.9691920510470589,0.705835675717647,0.5937042609803921}{\textbf{c}} \textcolor[rgb]{0.9120325752980393,0.469679582172549,0.36656490445882356}{\textbf{b}} \textcolor[rgb]{0.705673158,0.01555616,0.150232812}{\textbf{b}}  & \textcolor[rgb]{0.705673158,0.01555616,0.150232812}{\textbf{a}} \textcolor[rgb]{0.8674276350862745,0.864376599772549,0.8626024620196079}{\textbf{c}} \textcolor[rgb]{0.908908026654902,0.46243263716862765,0.36095039415294133}{\textbf{b}} \textcolor[rgb]{0.908908026654902,0.46243263716862765,0.36095039415294133}{\textbf{b}}  & \textcolor[rgb]{0.705673158,0.01555616,0.150232812}{\textbf{a}} \textcolor[rgb]{0.8674276350862745,0.864376599772549,0.8626024620196079}{\textbf{c}} \textcolor[rgb]{0.8836871397764705,0.8561077179529412,0.8402576701764708}{\textbf{b}} \textcolor[rgb]{0.8836871397764705,0.8561077179529412,0.8402576701764708}{\textbf{b}}  & \textcolor[rgb]{0.8674276350862745,0.864376599772549,0.8626024620196079}{\textbf{a}} \textcolor[rgb]{0.8674276350862745,0.864376599772549,0.8626024620196079}{\textbf{c}} \textcolor[rgb]{0.705673158,0.01555616,0.150232812}{\textbf{b}} \textcolor[rgb]{0.705673158,0.01555616,0.150232812}{\textbf{b}}  \\
		\heatmaprow{True}{True} \textcolor[rgb]{0.8714925112588235,0.8623093793176471,0.8570162640588236}{\textbf{a}} \textcolor[rgb]{0.8714925112588235,0.8623093793176471,0.8570162640588236}{\textbf{a}} \textcolor[rgb]{0.8674276350862745,0.864376599772549,0.8626024620196079}{\textbf{c}} \textcolor[rgb]{0.705673158,0.01555616,0.150232812}{\textbf{b}}  & \textcolor[rgb]{0.8674276350862745,0.864376599772549,0.8626024620196079}{\textbf{a}} \textcolor[rgb]{0.9593847296274509,0.6103057604117648,0.4893818509411765}{\textbf{a}} \textcolor[rgb]{0.9582793979254902,0.604335096882353,0.48329710138823534}{\textbf{c}} \textcolor[rgb]{0.705673158,0.01555616,0.150232812}{\textbf{b}} & \textcolor[rgb]{0.8674276350862745,0.864376599772549,0.8626024620196079}{\textbf{a}} \textcolor[rgb]{0.8674276350862745,0.864376599772549,0.8626024620196079}{\textbf{a}} \textcolor[rgb]{0.8674276350862745,0.864376599772549,0.8626024620196079}{\textbf{c}} \textcolor[rgb]{0.705673158,0.01555616,0.150232812}{\textbf{b}}  & \textcolor[rgb]{0.9604900613294117,0.6162764239411764,0.49546660049411767}{\textbf{a}} \textcolor[rgb]{0.9604900613294117,0.6162764239411764,0.49546660049411767}{\textbf{a}} \textcolor[rgb]{0.8674276350862745,0.864376599772549,0.8626024620196079}{\textbf{c}} \textcolor[rgb]{0.705673158,0.01555616,0.150232812}{\textbf{b}}  & \textcolor[rgb]{0.8674276350862745,0.864376599772549,0.8626024620196079}{\textbf{a}} \textcolor[rgb]{0.8674276350862745,0.864376599772549,0.8626024620196079}{\textbf{a}} \textcolor[rgb]{0.8674276350862745,0.864376599772549,0.8626024620196079}{\textbf{c}} \textcolor[rgb]{0.705673158,0.01555616,0.150232812}{\textbf{b}}  \\
		\heatmaprow{True}{True} \textcolor[rgb]{0.9695216017294117,0.7008328676235294,0.5875080175294117}{\textbf{a}} \textcolor[rgb]{0.705673158,0.01555616,0.150232812}{\textbf{b}} \textcolor[rgb]{0.9151571239411764,0.4769265271764706,0.3721794147647059}{\textbf{c}} \textcolor[rgb]{0.8674276350862745,0.864376599772549,0.8626024620196079}{\textbf{a}} \textcolor[rgb]{0.9695216017294117,0.7008328676235294,0.5875080175294117}{\textbf{b}}  & \textcolor[rgb]{0.8808963866470588,0.4023312782745098,0.3171151874901961}{\textbf{a}} \textcolor[rgb]{0.9151571239411764,0.4769265271764706,0.3721794147647059}{\textbf{b}} \textcolor[rgb]{0.9696829796666666,0.6904839307372549,0.5751383613647059}{\textbf{c}} \textcolor[rgb]{0.705673158,0.01555616,0.150232812}{\textbf{a}} \textcolor[rgb]{0.7409573187529412,0.12224032527058823,0.17574419910588235}{\textbf{b}}   & \textcolor[rgb]{0.9616447383764706,0.7580291825411765,0.6617823791647058}{\textbf{a}} \textcolor[rgb]{0.705673158,0.01555616,0.150232812}{\textbf{b}} \textcolor[rgb]{0.8155083866078432,0.2777809871764706,0.24029356566666665}{\textbf{c}} \textcolor[rgb]{0.8674276350862745,0.864376599772549,0.8626024620196079}{\textbf{a}} \textcolor[rgb]{0.9695216017294117,0.7008328676235294,0.5875080175294117}{\textbf{b}}  & \textcolor[rgb]{0.8204010983882353,0.2867649126352941,0.2451595198}{\textbf{a}} \textcolor[rgb]{0.705673158,0.01555616,0.150232812}{\textbf{b}} \textcolor[rgb]{0.9094595977529412,0.8393864797647058,0.8003313524235294}{\textbf{c}} \textcolor[rgb]{0.9595176584705882,0.7669728545098039,0.6741447150392157}{\textbf{a}} \textcolor[rgb]{0.9593847296274509,0.6103057604117648,0.4893818509411765}{\textbf{b}}  & \textcolor[rgb]{0.8674276350862745,0.864376599772549,0.8626024620196079}{\textbf{a}} \textcolor[rgb]{0.8523781350078431,0.34649194649411763,0.2803464686980392}{\textbf{b}} \textcolor[rgb]{0.705673158,0.01555616,0.150232812}{\textbf{c}} \textcolor[rgb]{0.8674276350862745,0.864376599772549,0.8626024620196079}{\textbf{a}} \textcolor[rgb]{0.8437026697215686,0.3300681294588235,0.2700648102196078}{\textbf{b}}  \\
		\heatmaprow{True}{False} \textcolor[rgb]{0.8674276350862745,0.864376599772549,0.8626024620196079}{\textbf{a}} \textcolor[rgb]{0.8674276350862745,0.864376599772549,0.8626024620196079}{\textbf{a}} \textcolor[rgb]{0.8674276350862745,0.864376599772549,0.8626024620196079}{\textbf{c}} \textcolor[rgb]{0.8674276350862745,0.864376599772549,0.8626024620196079}{\textbf{c}}  & \textcolor[rgb]{0.8674276350862745,0.864376599772549,0.8626024620196079}{\textbf{a}} \textcolor[rgb]{0.8674276350862745,0.864376599772549,0.8626024620196079}{\textbf{a}} \textcolor[rgb]{0.8674276350862745,0.864376599772549,0.8626024620196079}{\textbf{c}} \textcolor[rgb]{0.8674276350862745,0.864376599772549,0.8626024620196079}{\textbf{c}}  & \textcolor[rgb]{0.8674276350862745,0.864376599772549,0.8626024620196079}{\textbf{a}} \textcolor[rgb]{0.8674276350862745,0.864376599772549,0.8626024620196079}{\textbf{a}} \textcolor[rgb]{0.8674276350862745,0.864376599772549,0.8626024620196079}{\textbf{c}} \textcolor[rgb]{0.8674276350862745,0.864376599772549,0.8626024620196079}{\textbf{c}}  & \textcolor[rgb]{0.8674276350862745,0.864376599772549,0.8626024620196079}{\textbf{a}} \textcolor[rgb]{0.8674276350862745,0.864376599772549,0.8626024620196079}{\textbf{a}} \textcolor[rgb]{0.8674276350862745,0.864376599772549,0.8626024620196079}{\textbf{c}} \textcolor[rgb]{0.8674276350862745,0.864376599772549,0.8626024620196079}{\textbf{c}}  & \textcolor[rgb]{0.8674276350862745,0.864376599772549,0.8626024620196079}{\textbf{a}} \textcolor[rgb]{0.8674276350862745,0.864376599772549,0.8626024620196079}{\textbf{a}} \textcolor[rgb]{0.8674276350862745,0.864376599772549,0.8626024620196079}{\textbf{c}} \textcolor[rgb]{0.8674276350862745,0.864376599772549,0.8626024620196079}{\textbf{c}}  \\
		\heatmaprow{False}{True} \textcolor[rgb]{0.8674276350862745,0.864376599772549,0.8626024620196079}{\textbf{a}} \textcolor[rgb]{0.8674276350862745,0.864376599772549,0.8626024620196079}{\textbf{c}} \textcolor[rgb]{0.8674276350862745,0.864376599772549,0.8626024620196079}{\textbf{b}}  & \textcolor[rgb]{0.8674276350862745,0.864376599772549,0.8626024620196079}{\textbf{a}} \textcolor[rgb]{0.8674276350862745,0.864376599772549,0.8626024620196079}{\textbf{c}} \textcolor[rgb]{0.8674276350862745,0.864376599772549,0.8626024620196079}{\textbf{b}}  & \textcolor[rgb]{0.8674276350862745,0.864376599772549,0.8626024620196079}{\textbf{a}} \textcolor[rgb]{0.8674276350862745,0.864376599772549,0.8626024620196079}{\textbf{c}} \textcolor[rgb]{0.8674276350862745,0.864376599772549,0.8626024620196079}{\textbf{b}}  & \textcolor[rgb]{0.8674276350862745,0.864376599772549,0.8626024620196079}{\textbf{a}} \textcolor[rgb]{0.8674276350862745,0.864376599772549,0.8626024620196079}{\textbf{c}} \textcolor[rgb]{0.8674276350862745,0.864376599772549,0.8626024620196079}{\textbf{b}}  & \textcolor[rgb]{0.8674276350862745,0.864376599772549,0.8626024620196079}{\textbf{a}} \textcolor[rgb]{0.8674276350862745,0.864376599772549,0.8626024620196079}{\textbf{c}} \textcolor[rgb]{0.8674276350862745,0.864376599772549,0.8626024620196079}{\textbf{b}}  \\
		\heatmaprow{False}{False} \textcolor[rgb]{0.8674276350862745,0.864376599772549,0.8626024620196079}{\textbf{a}} \textcolor[rgb]{0.8674276350862745,0.864376599772549,0.8626024620196079}{\textbf{a}} \textcolor[rgb]{0.8674276350862745,0.864376599772549,0.8626024620196079}{\textbf{c}} \textcolor[rgb]{0.8674276350862745,0.864376599772549,0.8626024620196079}{\textbf{c}}  & \textcolor[rgb]{0.8674276350862745,0.864376599772549,0.8626024620196079}{\textbf{a}} \textcolor[rgb]{0.8674276350862745,0.864376599772549,0.8626024620196079}{\textbf{a}} \textcolor[rgb]{0.8674276350862745,0.864376599772549,0.8626024620196079}{\textbf{c}} \textcolor[rgb]{0.8674276350862745,0.864376599772549,0.8626024620196079}{\textbf{c}}  & \textcolor[rgb]{0.8674276350862745,0.864376599772549,0.8626024620196079}{\textbf{a}} \textcolor[rgb]{0.8674276350862745,0.864376599772549,0.8626024620196079}{\textbf{a}} \textcolor[rgb]{0.8674276350862745,0.864376599772549,0.8626024620196079}{\textbf{c}} \textcolor[rgb]{0.8674276350862745,0.864376599772549,0.8626024620196079}{\textbf{c}}  & \textcolor[rgb]{0.8674276350862745,0.864376599772549,0.8626024620196079}{\textbf{a}} \textcolor[rgb]{0.8674276350862745,0.864376599772549,0.8626024620196079}{\textbf{a}} \textcolor[rgb]{0.8674276350862745,0.864376599772549,0.8626024620196079}{\textbf{c}} \textcolor[rgb]{0.8674276350862745,0.864376599772549,0.8626024620196079}{\textbf{c}}  & \textcolor[rgb]{0.8674276350862745,0.864376599772549,0.8626024620196079}{\textbf{a}} \textcolor[rgb]{0.8674276350862745,0.864376599772549,0.8626024620196079}{\textbf{a}} \textcolor[rgb]{0.8674276350862745,0.864376599772549,0.8626024620196079}{\textbf{c}} \textcolor[rgb]{0.8674276350862745,0.864376599772549,0.8626024620196079}{\textbf{c}}  \\
		\hline
		
		\multirow{7}{*}{SP (FSA)} 
		\heatmaprow{True}{True} \textcolor[rgb]{0.705673158,0.01555616,0.150232812}{\textbf{a}} \textcolor[rgb]{0.705673158,0.01555616,0.150232812}{\textbf{c}} \textcolor[rgb]{0.705673158,0.01555616,0.150232812}{\textbf{b}}  & \textcolor[rgb]{0.863391831290196,0.8650837958196078,0.8676338842627451}{\textbf{a}} \textcolor[rgb]{0.8593850998705882,0.8644309674588235,0.8721105307882353}{\textbf{c}} \textcolor[rgb]{0.705673158,0.01555616,0.150232812}{\textbf{b}}  & \textcolor[rgb]{0.8674276350862745,0.864376599772549,0.8626024620196079}{\textbf{a}} \textcolor[rgb]{0.8674276350862745,0.864376599772549,0.8626024620196079}{\textbf{c}} \textcolor[rgb]{0.705673158,0.01555616,0.150232812}{\textbf{b}}  & \textcolor[rgb]{0.705673158,0.01555616,0.150232812}{\textbf{a}} \textcolor[rgb]{0.9695216017294117,0.7008328676235294,0.5875080175294117}{\textbf{c}} \textcolor[rgb]{0.9491505082901961,0.7907852690666667,0.7108755943019608}{\textbf{b}}  & \textcolor[rgb]{0.8674276350862745,0.864376599772549,0.8626024620196079}{\textbf{a}} \textcolor[rgb]{0.8674276350862745,0.864376599772549,0.8626024620196079}{\textbf{c}} \textcolor[rgb]{0.705673158,0.01555616,0.150232812}{\textbf{b}}  \\
		\heatmaprow{True}{True} \textcolor[rgb]{0.705673158,0.01555616,0.150232812}{\textbf{a}} \textcolor[rgb]{0.705673158,0.01555616,0.150232812}{\textbf{c}} \textcolor[rgb]{0.705673158,0.01555616,0.150232812}{\textbf{b}} \textcolor[rgb]{0.705673158,0.01555616,0.150232812}{\textbf{b}}  & \textcolor[rgb]{0.8674276350862745,0.864376599772549,0.8626024620196079}{\textbf{a}} \textcolor[rgb]{0.8674276350862745,0.864376599772549,0.8626024620196079}{\textbf{c}} \textcolor[rgb]{0.8674276350862745,0.864376599772549,0.8626024620196079}{\textbf{b}} \textcolor[rgb]{0.705673158,0.01555616,0.150232812}{\textbf{b}}  & \textcolor[rgb]{0.8674276350862745,0.864376599772549,0.8626024620196079}{\textbf{a}} \textcolor[rgb]{0.8674276350862745,0.864376599772549,0.8626024620196079}{\textbf{c}} \textcolor[rgb]{0.8674276350862745,0.864376599772549,0.8626024620196079}{\textbf{b}} \textcolor[rgb]{0.705673158,0.01555616,0.150232812}{\textbf{b}}  & \textcolor[rgb]{0.8252938101686275,0.2957488380941176,0.2500254739333333}{\textbf{a}} \textcolor[rgb]{0.705673158,0.01555616,0.150232812}{\textbf{c}} \textcolor[rgb]{0.9367796132117647,0.5327495001098039,0.41809333948627453}{\textbf{b}} \textcolor[rgb]{0.7233152383764706,0.06889824263529411,0.16298850555294117}{\textbf{b}}  & \textcolor[rgb]{0.8674276350862745,0.864376599772549,0.8626024620196079}{\textbf{a}} \textcolor[rgb]{0.8674276350862745,0.864376599772549,0.8626024620196079}{\textbf{c}} \textcolor[rgb]{0.8674276350862745,0.864376599772549,0.8626024620196079}{\textbf{b}} \textcolor[rgb]{0.705673158,0.01555616,0.150232812}{\textbf{b}}  \\
		\heatmaprow{True}{True} \textcolor[rgb]{0.8674276350862745,0.864376599772549,0.8626024620196079}{\textbf{a}} \textcolor[rgb]{0.705673158,0.01555616,0.150232812}{\textbf{a}} \textcolor[rgb]{0.705673158,0.01555616,0.150232812}{\textbf{c}} \textcolor[rgb]{0.705673158,0.01555616,0.150232812}{\textbf{b}}  & \textcolor[rgb]{0.8674276350862745,0.864376599772549,0.8626024620196079}{\textbf{a}} \textcolor[rgb]{0.8674276350862745,0.864376599772549,0.8626024620196079}{\textbf{a}} \textcolor[rgb]{0.8674276350862745,0.864376599772549,0.8626024620196079}{\textbf{c}} \textcolor[rgb]{0.705673158,0.01555616,0.150232812}{\textbf{b}}  & \textcolor[rgb]{0.8674276350862745,0.864376599772549,0.8626024620196079}{\textbf{a}} \textcolor[rgb]{0.8674276350862745,0.864376599772549,0.8626024620196079}{\textbf{a}} \textcolor[rgb]{0.8674276350862745,0.864376599772549,0.8626024620196079}{\textbf{c}} \textcolor[rgb]{0.705673158,0.01555616,0.150232812}{\textbf{b}}  & \textcolor[rgb]{0.8674276350862745,0.864376599772549,0.8626024620196079}{\textbf{a}} \textcolor[rgb]{0.705673158,0.01555616,0.150232812}{\textbf{a}} \textcolor[rgb]{0.9695216017294117,0.7008328676235294,0.5875080175294117}{\textbf{c}} \textcolor[rgb]{0.9491505082901961,0.7907852690666667,0.7108755943019608}{\textbf{b}}  & \textcolor[rgb]{0.8674276350862745,0.864376599772549,0.8626024620196079}{\textbf{a}} \textcolor[rgb]{0.8674276350862745,0.864376599772549,0.8626024620196079}{\textbf{a}} \textcolor[rgb]{0.8674276350862745,0.864376599772549,0.8626024620196079}{\textbf{c}} \textcolor[rgb]{0.705673158,0.01555616,0.150232812}{\textbf{b}}  \\
		\heatmaprow{True}{True} \textcolor[rgb]{0.8674276350862745,0.864376599772549,0.8626024620196079}{\textbf{a}} \textcolor[rgb]{0.8674276350862745,0.864376599772549,0.8626024620196079}{\textbf{b}} \textcolor[rgb]{0.8674276350862745,0.864376599772549,0.8626024620196079}{\textbf{c}} \textcolor[rgb]{0.705673158,0.01555616,0.150232812}{\textbf{a}} \textcolor[rgb]{0.705673158,0.01555616,0.150232812}{\textbf{b}}  & \textcolor[rgb]{0.8674276350862745,0.864376599772549,0.8626024620196079}{\textbf{a}} \textcolor[rgb]{0.8674276350862745,0.864376599772549,0.8626024620196079}{\textbf{b}} \textcolor[rgb]{0.863391831290196,0.8650837958196078,0.8676338842627451}{\textbf{c}} \textcolor[rgb]{0.8674276350862745,0.864376599772549,0.8626024620196079}{\textbf{a}} \textcolor[rgb]{0.705673158,0.01555616,0.150232812}{\textbf{b}}  & \textcolor[rgb]{0.8674276350862745,0.864376599772549,0.8626024620196079}{\textbf{a}} \textcolor[rgb]{0.8674276350862745,0.864376599772549,0.8626024620196079}{\textbf{b}} \textcolor[rgb]{0.863391831290196,0.8650837958196078,0.8676338842627451}{\textbf{c}} \textcolor[rgb]{0.8674276350862745,0.864376599772549,0.8626024620196079}{\textbf{a}} \textcolor[rgb]{0.705673158,0.01555616,0.150232812}{\textbf{b}}  & \textcolor[rgb]{0.29471843211764703,0.39354192974117647,0.8343841671215686}{\textbf{a}} \textcolor[rgb]{0.705673158,0.01555616,0.150232812}{\textbf{b}} \textcolor[rgb]{0.9649113881372549,0.6401590780588234,0.5198055987058824}{\textbf{c}} \textcolor[rgb]{0.9434315296666667,0.8022762536156862,0.7291715979137255}{\textbf{a}} \textcolor[rgb]{0.9616447383764706,0.7580291825411765,0.6617823791647058}{\textbf{b}}  & \textcolor[rgb]{0.8674276350862745,0.864376599772549,0.8626024620196079}{\textbf{a}} \textcolor[rgb]{0.8674276350862745,0.864376599772549,0.8626024620196079}{\textbf{b}} \textcolor[rgb]{0.8674276350862745,0.864376599772549,0.8626024620196079}{\textbf{c}} \textcolor[rgb]{0.8674276350862745,0.864376599772549,0.8626024620196079}{\textbf{a}} \textcolor[rgb]{0.705673158,0.01555616,0.150232812}{\textbf{b}}  \\
		\heatmaprow{True}{False} \textcolor[rgb]{0.8674276350862745,0.864376599772549,0.8626024620196079}{\textbf{a}} \textcolor[rgb]{0.8674276350862745,0.864376599772549,0.8626024620196079}{\textbf{a}} \textcolor[rgb]{0.8674276350862745,0.864376599772549,0.8626024620196079}{\textbf{c}} \textcolor[rgb]{0.8674276350862745,0.864376599772549,0.8626024620196079}{\textbf{c}}  & \textcolor[rgb]{0.8674276350862745,0.864376599772549,0.8626024620196079}{\textbf{a}} \textcolor[rgb]{0.8674276350862745,0.864376599772549,0.8626024620196079}{\textbf{a}} \textcolor[rgb]{0.8674276350862745,0.864376599772549,0.8626024620196079}{\textbf{c}} \textcolor[rgb]{0.8674276350862745,0.864376599772549,0.8626024620196079}{\textbf{c}}  & \textcolor[rgb]{0.8674276350862745,0.864376599772549,0.8626024620196079}{\textbf{a}} \textcolor[rgb]{0.8674276350862745,0.864376599772549,0.8626024620196079}{\textbf{a}} \textcolor[rgb]{0.8674276350862745,0.864376599772549,0.8626024620196079}{\textbf{c}} \textcolor[rgb]{0.8674276350862745,0.864376599772549,0.8626024620196079}{\textbf{c}}  & \textcolor[rgb]{0.8674276350862745,0.864376599772549,0.8626024620196079}{\textbf{a}} \textcolor[rgb]{0.8674276350862745,0.864376599772549,0.8626024620196079}{\textbf{a}} \textcolor[rgb]{0.8674276350862745,0.864376599772549,0.8626024620196079}{\textbf{c}} \textcolor[rgb]{0.8674276350862745,0.864376599772549,0.8626024620196079}{\textbf{c}}  & \textcolor[rgb]{0.8674276350862745,0.864376599772549,0.8626024620196079}{\textbf{a}} \textcolor[rgb]{0.8674276350862745,0.864376599772549,0.8626024620196079}{\textbf{a}} \textcolor[rgb]{0.8674276350862745,0.864376599772549,0.8626024620196079}{\textbf{c}} \textcolor[rgb]{0.8674276350862745,0.864376599772549,0.8626024620196079}{\textbf{c}}  \\
		\heatmaprow{False}{True} \textcolor[rgb]{0.2298057,0.298717966,0.753683153}{\textbf{a}} \textcolor[rgb]{0.2298057,0.298717966,0.753683153}{\textbf{c}} \textcolor[rgb]{0.8674276350862745,0.864376599772549,0.8626024620196079}{\textbf{b}}  & \textcolor[rgb]{0.8674276350862745,0.864376599772549,0.8626024620196079}{\textbf{a}} \textcolor[rgb]{0.8674276350862745,0.864376599772549,0.8626024620196079}{\textbf{c}} \textcolor[rgb]{0.8674276350862745,0.864376599772549,0.8626024620196079}{\textbf{b}}  & \textcolor[rgb]{0.8674276350862745,0.864376599772549,0.8626024620196079}{\textbf{a}} \textcolor[rgb]{0.8674276350862745,0.864376599772549,0.8626024620196079}{\textbf{c}} \textcolor[rgb]{0.8674276350862745,0.864376599772549,0.8626024620196079}{\textbf{b}}  & \textcolor[rgb]{0.6933212848235294,0.7963141317058823,0.9863077805294118}{\textbf{a}} \textcolor[rgb]{0.5054226428235293,0.6439946735686275,0.9831574312745098}{\textbf{c}} \textcolor[rgb]{0.705673158,0.01555616,0.150232812}{\textbf{b}}  & \textcolor[rgb]{0.8674276350862745,0.864376599772549,0.8626024620196079}{\textbf{a}} \textcolor[rgb]{0.8674276350862745,0.864376599772549,0.8626024620196079}{\textbf{c}} \textcolor[rgb]{0.8674276350862745,0.864376599772549,0.8626024620196079}{\textbf{b}}  \\
		\heatmaprow{False}{False} \textcolor[rgb]{0.8674276350862745,0.864376599772549,0.8626024620196079}{\textbf{a}} \textcolor[rgb]{0.8674276350862745,0.864376599772549,0.8626024620196079}{\textbf{a}} \textcolor[rgb]{0.8674276350862745,0.864376599772549,0.8626024620196079}{\textbf{c}} \textcolor[rgb]{0.705673158,0.01555616,0.150232812}{\textbf{c}}  & \textcolor[rgb]{0.8674276350862745,0.864376599772549,0.8626024620196079}{\textbf{a}} \textcolor[rgb]{0.863391831290196,0.8650837958196078,0.8676338842627451}{\textbf{a}} \textcolor[rgb]{0.863391831290196,0.8650837958196078,0.8676338842627451}{\textbf{c}} \textcolor[rgb]{0.705673158,0.01555616,0.150232812}{\textbf{c}}  & \textcolor[rgb]{0.8674276350862745,0.864376599772549,0.8626024620196079}{\textbf{a}} \textcolor[rgb]{0.863391831290196,0.8650837958196078,0.8676338842627451}{\textbf{a}} \textcolor[rgb]{0.863391831290196,0.8650837958196078,0.8676338842627451}{\textbf{c}} \textcolor[rgb]{0.705673158,0.01555616,0.150232812}{\textbf{c}}  & \textcolor[rgb]{0.8674276350862745,0.864376599772549,0.8626024620196079}{\textbf{a}} \textcolor[rgb]{0.8958817682941177,0.8499060565882353,0.8234990762941177}{\textbf{a}} \textcolor[rgb]{0.8674276350862745,0.864376599772549,0.8626024620196079}{\textbf{c}} \textcolor[rgb]{0.705673158,0.01555616,0.150232812}{\textbf{c}}  & \textcolor[rgb]{0.8674276350862745,0.864376599772549,0.8626024620196079}{\textbf{a}} \textcolor[rgb]{0.8674276350862745,0.864376599772549,0.8626024620196079}{\textbf{a}} \textcolor[rgb]{0.8674276350862745,0.864376599772549,0.8626024620196079}{\textbf{c}} \textcolor[rgb]{0.705673158,0.01555616,0.150232812}{\textbf{c}}  \\
		\hline
		
		\multirow{5}{*}{Bracket (PDA)}
		\heatmaprow{]}{]} \textcolor[rgb]{0.8674276350862745,0.864376599772549,0.8626024620196079}{\textbf{(}} \textcolor[rgb]{0.8674276350862745,0.864376599772549,0.8626024620196079}{\textbf{[}} \textcolor[rgb]{0.8674276350862745,0.864376599772549,0.8626024620196079}{\textbf{[}} \textcolor[rgb]{0.8674276350862745,0.864376599772549,0.8626024620196079}{\textbf{(}} \textcolor[rgb]{0.8674276350862745,0.864376599772549,0.8626024620196079}{\textbf{[}}  & \textcolor[rgb]{0.8674276350862745,0.864376599772549,0.8626024620196079}{\textbf{(}} \textcolor[rgb]{0.8674276350862745,0.864376599772549,0.8626024620196079}{\textbf{[}} \textcolor[rgb]{0.8674276350862745,0.864376599772549,0.8626024620196079}{\textbf{[}} \textcolor[rgb]{0.8674276350862745,0.864376599772549,0.8626024620196079}{\textbf{(}} \textcolor[rgb]{0.8674276350862745,0.864376599772549,0.8626024620196079}{\textbf{[}}  & \textcolor[rgb]{0.8674276350862745,0.864376599772549,0.8626024620196079}{\textbf{(}} \textcolor[rgb]{0.8674276350862745,0.864376599772549,0.8626024620196079}{\textbf{[}} \textcolor[rgb]{0.8674276350862745,0.864376599772549,0.8626024620196079}{\textbf{[}} \textcolor[rgb]{0.8674276350862745,0.864376599772549,0.8626024620196079}{\textbf{(}} \textcolor[rgb]{0.8674276350862745,0.864376599772549,0.8626024620196079}{\textbf{[}}  & \textcolor[rgb]{0.8674276350862745,0.864376599772549,0.8626024620196079}{\textbf{(}} \textcolor[rgb]{0.8674276350862745,0.864376599772549,0.8626024620196079}{\textbf{[}} \textcolor[rgb]{0.8674276350862745,0.864376599772549,0.8626024620196079}{\textbf{[}} \textcolor[rgb]{0.8674276350862745,0.864376599772549,0.8626024620196079}{\textbf{(}} \textcolor[rgb]{0.8674276350862745,0.864376599772549,0.8626024620196079}{\textbf{[}}  & \textcolor[rgb]{0.8674276350862745,0.864376599772549,0.8626024620196079}{\textbf{(}} \textcolor[rgb]{0.8674276350862745,0.864376599772549,0.8626024620196079}{\textbf{[}} \textcolor[rgb]{0.8674276350862745,0.864376599772549,0.8626024620196079}{\textbf{[}} \textcolor[rgb]{0.8674276350862745,0.864376599772549,0.8626024620196079}{\textbf{(}} \textcolor[rgb]{0.705673158,0.01555616,0.150232812}{\textbf{[}} \\
		\heatmaprow{)}{)} \textcolor[rgb]{0.8674276350862745,0.864376599772549,0.8626024620196079}{\textbf{(}} \textcolor[rgb]{0.8674276350862745,0.864376599772549,0.8626024620196079}{\textbf{[}} \textcolor[rgb]{0.8674276350862745,0.864376599772549,0.8626024620196079}{\textbf{[}} \textcolor[rgb]{0.705673158,0.01555616,0.150232812}{\textbf{(}} \textcolor[rgb]{0.8674276350862745,0.864376599772549,0.8626024620196079}{\textbf{[}} \textcolor[rgb]{0.8674276350862745,0.864376599772549,0.8626024620196079}{\textbf{]}}  & \textcolor[rgb]{0.8674276350862745,0.864376599772549,0.8626024620196079}{\textbf{(}} \textcolor[rgb]{0.8674276350862745,0.864376599772549,0.8626024620196079}{\textbf{[}} \textcolor[rgb]{0.8674276350862745,0.864376599772549,0.8626024620196079}{\textbf{[}} \textcolor[rgb]{0.8674276350862745,0.864376599772549,0.8626024620196079}{\textbf{(}} \textcolor[rgb]{0.8674276350862745,0.864376599772549,0.8626024620196079}{\textbf{[}} \textcolor[rgb]{0.8674276350862745,0.864376599772549,0.8626024620196079}{\textbf{]}}  & \textcolor[rgb]{0.8674276350862745,0.864376599772549,0.8626024620196079}{\textbf{(}} \textcolor[rgb]{0.8674276350862745,0.864376599772549,0.8626024620196079}{\textbf{[}} \textcolor[rgb]{0.8674276350862745,0.864376599772549,0.8626024620196079}{\textbf{[}} \textcolor[rgb]{0.8674276350862745,0.864376599772549,0.8626024620196079}{\textbf{(}} \textcolor[rgb]{0.8674276350862745,0.864376599772549,0.8626024620196079}{\textbf{[}} \textcolor[rgb]{0.8674276350862745,0.864376599772549,0.8626024620196079}{\textbf{]}}  & \textcolor[rgb]{0.8674276350862745,0.864376599772549,0.8626024620196079}{\textbf{(}} \textcolor[rgb]{0.8674276350862745,0.864376599772549,0.8626024620196079}{\textbf{[}} \textcolor[rgb]{0.8674276350862745,0.864376599772549,0.8626024620196079}{\textbf{[}} \textcolor[rgb]{0.8674276350862745,0.864376599772549,0.8626024620196079}{\textbf{(}} \textcolor[rgb]{0.8674276350862745,0.864376599772549,0.8626024620196079}{\textbf{[}} \textcolor[rgb]{0.8674276350862745,0.864376599772549,0.8626024620196079}{\textbf{]}}  & \textcolor[rgb]{0.9226814526235294,0.8285681381176471,0.7770543897882353}{\textbf{(}} \textcolor[rgb]{0.724041371882353,0.8149103926470588,0.9756509706470589}{\textbf{[}} \textcolor[rgb]{0.5543118699137254,0.6900970112156862,0.9955155482352941}{\textbf{[}} \textcolor[rgb]{0.705673158,0.01555616,0.150232812}{\textbf{(}} \textcolor[rgb]{0.8674276350862745,0.864376599772549,0.8626024620196079}{\textbf{[}} \textcolor[rgb]{0.8674276350862745,0.864376599772549,0.8626024620196079}{\textbf{]}}  \\
		\heatmaprow{None}{None} \textcolor[rgb]{0.8674276350862745,0.864376599772549,0.8626024620196079}{\textbf{(}} \textcolor[rgb]{0.8674276350862745,0.864376599772549,0.8626024620196079}{\textbf{[}} \textcolor[rgb]{0.8674276350862745,0.864376599772549,0.8626024620196079}{\textbf{[}} \textcolor[rgb]{0.8674276350862745,0.864376599772549,0.8626024620196079}{\textbf{]}} \textcolor[rgb]{0.8674276350862745,0.864376599772549,0.8626024620196079}{\textbf{]}} \textcolor[rgb]{0.8674276350862745,0.864376599772549,0.8626024620196079}{\textbf{)}}  & \textcolor[rgb]{0.8674276350862745,0.864376599772549,0.8626024620196079}{\textbf{(}} \textcolor[rgb]{0.8674276350862745,0.864376599772549,0.8626024620196079}{\textbf{[}} \textcolor[rgb]{0.8674276350862745,0.864376599772549,0.8626024620196079}{\textbf{[}} \textcolor[rgb]{0.8674276350862745,0.864376599772549,0.8626024620196079}{\textbf{]}} \textcolor[rgb]{0.8674276350862745,0.864376599772549,0.8626024620196079}{\textbf{]}} \textcolor[rgb]{0.8674276350862745,0.864376599772549,0.8626024620196079}{\textbf{)}}  & \textcolor[rgb]{0.8674276350862745,0.864376599772549,0.8626024620196079}{\textbf{(}} \textcolor[rgb]{0.8674276350862745,0.864376599772549,0.8626024620196079}{\textbf{[}} \textcolor[rgb]{0.8674276350862745,0.864376599772549,0.8626024620196079}{\textbf{[}} \textcolor[rgb]{0.8674276350862745,0.864376599772549,0.8626024620196079}{\textbf{]}} \textcolor[rgb]{0.8674276350862745,0.864376599772549,0.8626024620196079}{\textbf{]}} \textcolor[rgb]{0.8674276350862745,0.864376599772549,0.8626024620196079}{\textbf{)}}  & \textcolor[rgb]{0.8674276350862745,0.864376599772549,0.8626024620196079}{\textbf{(}} \textcolor[rgb]{0.8674276350862745,0.864376599772549,0.8626024620196079}{\textbf{[}} \textcolor[rgb]{0.8674276350862745,0.864376599772549,0.8626024620196079}{\textbf{[}} \textcolor[rgb]{0.8674276350862745,0.864376599772549,0.8626024620196079}{\textbf{]}} \textcolor[rgb]{0.8674276350862745,0.864376599772549,0.8626024620196079}{\textbf{]}} \textcolor[rgb]{0.8674276350862745,0.864376599772549,0.8626024620196079}{\textbf{)}}  & \textcolor[rgb]{0.2298057,0.298717966,0.753683153}{\textbf{(}} \textcolor[rgb]{0.2298057,0.298717966,0.753683153}{\textbf{[}} \textcolor[rgb]{0.2298057,0.298717966,0.753683153}{\textbf{[}} \textcolor[rgb]{0.8674276350862745,0.864376599772549,0.8626024620196079}{\textbf{]}} \textcolor[rgb]{0.8674276350862745,0.864376599772549,0.8626024620196079}{\textbf{]}} \textcolor[rgb]{0.8674276350862745,0.864376599772549,0.8626024620196079}{\textbf{)}}  \\
		\heatmaprow{]}{]} \textcolor[rgb]{0.8674276350862745,0.864376599772549,0.8626024620196079}{\textbf{[}} \textcolor[rgb]{0.8674276350862745,0.864376599772549,0.8626024620196079}{\textbf{(}} \textcolor[rgb]{0.705673158,0.01555616,0.150232812}{\textbf{[}} \textcolor[rgb]{0.8674276350862745,0.864376599772549,0.8626024620196079}{\textbf{]}} \textcolor[rgb]{0.705673158,0.01555616,0.150232812}{\textbf{[}} \textcolor[rgb]{0.8674276350862745,0.864376599772549,0.8626024620196079}{\textbf{(}} \textcolor[rgb]{0.8674276350862745,0.864376599772549,0.8626024620196079}{\textbf{)}}  & \textcolor[rgb]{0.8674276350862745,0.864376599772549,0.8626024620196079}{\textbf{[}} \textcolor[rgb]{0.8674276350862745,0.864376599772549,0.8626024620196079}{\textbf{(}} \textcolor[rgb]{0.8674276350862745,0.864376599772549,0.8626024620196079}{\textbf{[}} \textcolor[rgb]{0.8674276350862745,0.864376599772549,0.8626024620196079}{\textbf{]}} \textcolor[rgb]{0.8674276350862745,0.864376599772549,0.8626024620196079}{\textbf{[}} \textcolor[rgb]{0.8674276350862745,0.864376599772549,0.8626024620196079}{\textbf{(}} \textcolor[rgb]{0.8674276350862745,0.864376599772549,0.8626024620196079}{\textbf{)}}  & \textcolor[rgb]{0.8674276350862745,0.864376599772549,0.8626024620196079}{\textbf{[}} \textcolor[rgb]{0.8674276350862745,0.864376599772549,0.8626024620196079}{\textbf{(}} \textcolor[rgb]{0.8674276350862745,0.864376599772549,0.8626024620196079}{\textbf{[}} \textcolor[rgb]{0.8674276350862745,0.864376599772549,0.8626024620196079}{\textbf{]}} \textcolor[rgb]{0.8674276350862745,0.864376599772549,0.8626024620196079}{\textbf{[}} \textcolor[rgb]{0.8674276350862745,0.864376599772549,0.8626024620196079}{\textbf{(}} \textcolor[rgb]{0.8674276350862745,0.864376599772549,0.8626024620196079}{\textbf{)}}  & \textcolor[rgb]{0.8674276350862745,0.864376599772549,0.8626024620196079}{\textbf{[}} \textcolor[rgb]{0.8674276350862745,0.864376599772549,0.8626024620196079}{\textbf{(}} \textcolor[rgb]{0.8674276350862745,0.864376599772549,0.8626024620196079}{\textbf{[}} \textcolor[rgb]{0.8674276350862745,0.864376599772549,0.8626024620196079}{\textbf{]}} \textcolor[rgb]{0.8674276350862745,0.864376599772549,0.8626024620196079}{\textbf{[}} \textcolor[rgb]{0.8674276350862745,0.864376599772549,0.8626024620196079}{\textbf{(}} \textcolor[rgb]{0.8674276350862745,0.864376599772549,0.8626024620196079}{\textbf{)}}  & \textcolor[rgb]{0.8696552305058823,0.37927381945098043,0.30094110221960785}{\textbf{[}} \textcolor[rgb]{0.2298057,0.298717966,0.753683153}{\textbf{(}} \textcolor[rgb]{0.8674276350862745,0.864376599772549,0.8626024620196079}{\textbf{[}} \textcolor[rgb]{0.8674276350862745,0.864376599772549,0.8626024620196079}{\textbf{]}} \textcolor[rgb]{0.705673158,0.01555616,0.150232812}{\textbf{[}} \textcolor[rgb]{0.8674276350862745,0.864376599772549,0.8626024620196079}{\textbf{(}} \textcolor[rgb]{0.8674276350862745,0.864376599772549,0.8626024620196079}{\textbf{)}}  \\
		\heatmaprow{)}{]} \textcolor[rgb]{0.8674276350862745,0.864376599772549,0.8626024620196079}{\textbf{[}} \textcolor[rgb]{0.8674276350862745,0.864376599772549,0.8626024620196079}{\textbf{(}} \textcolor[rgb]{0.2298057,0.298717966,0.753683153}{\textbf{[}} \textcolor[rgb]{0.8674276350862745,0.864376599772549,0.8626024620196079}{\textbf{]}} \textcolor[rgb]{0.2298057,0.298717966,0.753683153}{\textbf{[}} \textcolor[rgb]{0.8674276350862745,0.864376599772549,0.8626024620196079}{\textbf{(}} \textcolor[rgb]{0.8674276350862745,0.864376599772549,0.8626024620196079}{\textbf{)}}  & \textcolor[rgb]{0.8674276350862745,0.864376599772549,0.8626024620196079}{\textbf{[}} \textcolor[rgb]{0.8674276350862745,0.864376599772549,0.8626024620196079}{\textbf{(}} \textcolor[rgb]{0.8674276350862745,0.864376599772549,0.8626024620196079}{\textbf{[}} \textcolor[rgb]{0.8674276350862745,0.864376599772549,0.8626024620196079}{\textbf{]}} \textcolor[rgb]{0.8674276350862745,0.864376599772549,0.8626024620196079}{\textbf{[}} \textcolor[rgb]{0.8674276350862745,0.864376599772549,0.8626024620196079}{\textbf{(}} \textcolor[rgb]{0.8674276350862745,0.864376599772549,0.8626024620196079}{\textbf{)}}  & \textcolor[rgb]{0.8674276350862745,0.864376599772549,0.8626024620196079}{\textbf{[}} \textcolor[rgb]{0.8674276350862745,0.864376599772549,0.8626024620196079}{\textbf{(}} \textcolor[rgb]{0.8674276350862745,0.864376599772549,0.8626024620196079}{\textbf{[}} \textcolor[rgb]{0.8674276350862745,0.864376599772549,0.8626024620196079}{\textbf{]}} \textcolor[rgb]{0.8674276350862745,0.864376599772549,0.8626024620196079}{\textbf{[}} \textcolor[rgb]{0.8674276350862745,0.864376599772549,0.8626024620196079}{\textbf{(}} \textcolor[rgb]{0.8674276350862745,0.864376599772549,0.8626024620196079}{\textbf{)}}  & \textcolor[rgb]{0.8674276350862745,0.864376599772549,0.8626024620196079}{\textbf{[}} \textcolor[rgb]{0.8674276350862745,0.864376599772549,0.8626024620196079}{\textbf{(}} \textcolor[rgb]{0.8674276350862745,0.864376599772549,0.8626024620196079}{\textbf{[}} \textcolor[rgb]{0.8674276350862745,0.864376599772549,0.8626024620196079}{\textbf{]}} \textcolor[rgb]{0.8674276350862745,0.864376599772549,0.8626024620196079}{\textbf{[}} \textcolor[rgb]{0.8674276350862745,0.864376599772549,0.8626024620196079}{\textbf{(}} \textcolor[rgb]{0.8674276350862745,0.864376599772549,0.8626024620196079}{\textbf{)}}  & \textcolor[rgb]{0.383662065772549,0.5101834172862746,0.9178306732313726}{\textbf{[}} \textcolor[rgb]{0.705673158,0.01555616,0.150232812}{\textbf{(}} \textcolor[rgb]{0.8674276350862745,0.864376599772549,0.8626024620196079}{\textbf{[}} \textcolor[rgb]{0.8674276350862745,0.864376599772549,0.8626024620196079}{\textbf{]}} \textcolor[rgb]{0.2298057,0.298717966,0.753683153}{\textbf{[}} \textcolor[rgb]{0.8674276350862745,0.864376599772549,0.8626024620196079}{\textbf{(}} \textcolor[rgb]{0.8674276350862745,0.864376599772549,0.8626024620196079}{\textbf{)}}  \\\hline
		
	\end{tabular}
	\caption{Selected heatmaps based on $R^{(c)}_t(\bm{X})$. \textcolor[rgb]{0.705673158,0.01555616,0.150232812}{\textbf{Red}} represents positive values and \textcolor[rgb]{0.2298057,0.298717966,0.753683153}{\textbf{blue}} represents negative values. Heatmaps with all values within the range of $\pm 1 \times 10^{-5}$ are shown as all $0$s.}
	\label{fig:heatmaps}
\end{table*}

To evaluate attribution methods under our framework, we begin with a qualitative description of the heatmaps that are computed for our white-box networks, based on the illustrative sample of heatmaps appearing in  \autoref{fig:heatmaps}. 

\subsection{Counting Task}

Occlusion, G $\times$ I, and IG are well-behaved for the counting task. As expected, these methods assign \ta\ a positive value and \tb\ a negative value when the output class for attribution is $c = \textit{True}$. When the number of \ta s is different from the number of \tb s, occlusion assigns a lower-magnitude score to the symbol with fewer instances. When $c = \textit{False}$, all relevance scores are $0$. This is because $\hat{y}_{\textit{False}}$ is fixed to a constant value supplied by a bias term, so input features cannot affect its value. 

Saliency and LRP both fail to produce nonzero scores, at least in some cases. Saliency scores satisfy $R^{(\textit{True})}_{t, 1}(\bm{X}) = -R^{(\textit{True})}_{t, 2}(\bm{X})$, resulting in token-level scores of $0$ for all inputs. Heatmaps \#3 and \#4 show that LRP assigns scores of $0$ to prefixes containing equal numbers of \ta s and \tb s. We will see in \autoref{sec:saturation} that this phenomenon appears to be related to the fact that the LSTM gates are saturated.

\subsection{SP Task}

We obtain radically different heatmaps for the two SP task networks, despite the fact that they produce the same classifications for all inputs.

For the counter-based network, all methods except for saliency assign positive scores for $c = \textit{True}$ to symbols constituting one of the four subsequences, and scores of zero elsewhere. The saliency heatmaps do not adhere to this pattern, and instead generally assign higher scores to tokens occurring near the end of the input. Heatmaps \#7--10 show that LRP fails to assign positive scores to the first symbol of each subsequence, while the other methods generally do not.\footnote{Although it is difficult to see, IG assigns a small positive score to the \tb s in heatmaps \#7 and \#8.} The LRP behavior reflects the fact that the initial \ta\ does not increment the subsequence counters, which determine the final logit score. In contrast, the behavior of occlusion, G $\times$ I, and IG is explained by the fact that removing either the \ta\ or the \tb\ destroys the subsequence. Note that the \ta s in heatmap \#9 receive scores of $0$ from occlusion and G $\times$ I, since removing only one of the two \ta s does not destroy the subsequence. 

For the FSA-based network, saliency, G $\times$ I, and LRP assign only the last symbol a nonzero score when the relevance output class $c$ matches the network's predicted class. IG appears to produce erratic heatmaps, exhibiting no immediately obvious pattern. Although occlusion appears to be erratic at first glance, its behavior can be explained by the fact that changing $\mathbf{x}^{(t)}$ to $\mathbf{0}$ causes $\mathbf{h}^{(t)}$ to be $\mathbf{0}$, which the LSTM interprets as the initial state of the FSA; thus, $R^{(c)}_t(\mathbf{X}) \neq 0$ precisely when $\mathbf{X}_{t + 1:, :}$ is classified differently from $\mathbf{X}$. In all cases, the heatmaps for the FSA-based network diverge significantly from the expected heatmaps.

\subsection{Bracket Prediction Task}

The heatmaps for the PDA-based network also differ strikingly from those of the other networks, in that the gradient-based methods never assign nonzero scores. This is because \autoref{eqn:bracketnetwork} causes $\bm{g}^{(t)}$ to be highly saturated, resulting in zero gradients. In the case of LRP, the matching bracket is highlighted when $c \neq \textit{None}$. When the matching bracket is not the last symbol of the input, the other unclosed brackets are also highlighted, with progressively smaller magnitudes, and with brackets of the opposite type from $c$ receiving negative scores. This pattern reflects the mechanism of (\ref{eqn:bracketnetwork}), in which progressively larger powers of $2$ are used to determine the content copied to $c_k^{(t)}$. When the relevance output class is $c = \textit{None}$, LRP assigns opening brackets a negative score, revealing the fact that those input symbols set the bit $c_{2k + 1}^{(t)}$ to indicate that the stack is not empty. Although occlusion sometimes highlights the matching bracket, it does not appear to be consistent in doing so. For example, it fails to highlight the matching bracket in heatmap \#21, and highlights one other bracket in heatmaps \#23--24.

\section{Detailed Evaluations}
\label{sec:detailedevaluations}

We now turn to focused investigations of particular phenomena that attribution methods exhibit when applied to white-box networks. \autoref{sec:saturation} begins by discussing the effect of network saturation on the gradient-based methods and LRP. In \autoref{sec:ablation} we apply  \citeauthor{bachPixelWiseExplanationsNonLinear2015}'s (\citeyear{bachPixelWiseExplanationsNonLinear2015}) ablation test to our attribution methods for the SP task.

\subsection{Saturation}
\label{sec:saturation}

\begin{table}
	\centering
	\small
	\begin{tabular}{c c c  l l l}
		\hline
		$u$ & $v$ & $\hat{y}_{\textit{True}}$  & \textbf{Saliency} & \textbf{G $\times$ I} & \textbf{IG} \\\hline
		0.6 & 0.537 & 0.151 &  \textcolor[rgb]{0.9057834780117647,0.4551856921647059,0.35533588384705883}{\textbf{a}} \textcolor[rgb]{0.7291959318352941,0.08667893684705881,0.16724040340392157}{\textbf{c}} \textcolor[rgb]{0.705673158,0.01555616,0.150232812}{\textbf{c}} \textcolor[rgb]{0.9281160096666666,0.8221971488627451,0.765141349254902}{\textbf{b}}  & \textcolor[rgb]{0.705673158,0.01555616,0.150232812}{\textbf{a}} \textcolor[rgb]{0.8674276350862745,0.864376599772549,0.8626024620196079}{\textbf{c}} \textcolor[rgb]{0.8674276350862745,0.864376599772549,0.8626024620196079}{\textbf{c}} \textcolor[rgb]{0.8836871397764705,0.8561077179529412,0.8402576701764708}{\textbf{b}}  & \textcolor[rgb]{0.705673158,0.01555616,0.150232812}{\textbf{a}} \textcolor[rgb]{0.8674276350862745,0.864376599772549,0.8626024620196079}{\textbf{c}} \textcolor[rgb]{0.8674276350862745,0.864376599772549,0.8626024620196079}{\textbf{c}} \textcolor[rgb]{0.8796222636039216,0.8581749384078431,0.845843868137255}{\textbf{b}}  \\
		0.7 & 0.604 & 0.533 &  \textcolor[rgb]{0.933221183,0.8155568504470588,0.7531514321411764}{\textbf{a}} \textcolor[rgb]{0.9440545734235294,0.5531534787490197,0.4355484903137255}{\textbf{c}} \textcolor[rgb]{0.705673158,0.01555616,0.150232812}{\textbf{c}} \textcolor[rgb]{0.9120325752980393,0.469679582172549,0.36656490445882356}{\textbf{b}}  & \textcolor[rgb]{0.8610536002941176,0.3629157635294118,0.2906281271764706}{\textbf{a}} \textcolor[rgb]{0.8674276350862745,0.864376599772549,0.8626024620196079}{\textbf{c}} \textcolor[rgb]{0.8674276350862745,0.864376599772549,0.8626024620196079}{\textbf{c}} \textcolor[rgb]{0.705673158,0.01555616,0.150232812}{\textbf{b}}  & \textcolor[rgb]{0.705673158,0.01555616,0.150232812}{\textbf{a}} \textcolor[rgb]{0.8674276350862745,0.864376599772549,0.8626024620196079}{\textbf{c}} \textcolor[rgb]{0.8674276350862745,0.864376599772549,0.8626024620196079}{\textbf{c}} \textcolor[rgb]{0.9094595977529412,0.8393864797647058,0.8003313524235294}{\textbf{b}}  \\
		0.8 & 0.664 & 0.581 &  \textcolor[rgb]{0.8674276350862745,0.864376599772549,0.8626024620196079}{\textbf{a}} \textcolor[rgb]{0.9684997476666667,0.673977379772549,0.5566492560470588}{\textbf{c}} \textcolor[rgb]{0.705673158,0.01555616,0.150232812}{\textbf{c}} \textcolor[rgb]{0.8301865219490197,0.30473276355294115,0.25489142806666665}{\textbf{b}}   & \textcolor[rgb]{0.8714925112588235,0.8623093793176471,0.8570162640588236}{\textbf{a}} \textcolor[rgb]{0.8674276350862745,0.864376599772549,0.8626024620196079}{\textbf{c}} \textcolor[rgb]{0.8674276350862745,0.864376599772549,0.8626024620196079}{\textbf{c}} \textcolor[rgb]{0.705673158,0.01555616,0.150232812}{\textbf{b}}  & \textcolor[rgb]{0.705673158,0.01555616,0.150232812}{\textbf{a}} \textcolor[rgb]{0.8674276350862745,0.864376599772549,0.8626024620196079}{\textbf{c}} \textcolor[rgb]{0.8674276350862745,0.864376599772549,0.8626024620196079}{\textbf{c}} \textcolor[rgb]{0.940878943,0.8055964028235294,0.7351665564705883}{\textbf{b}}  \\
		1 & 0.762 & 0.642 &  \textcolor[rgb]{0.8674276350862745,0.864376599772549,0.8626024620196079}{\textbf{a}} \textcolor[rgb]{0.9698511524117647,0.6958300595294117,0.5813117740784314}{\textbf{c}} \textcolor[rgb]{0.705673158,0.01555616,0.150232812}{\textbf{c}} \textcolor[rgb]{0.848040402364706,0.3382800379764708,0.2752056394588237}{\textbf{b}}  & \textcolor[rgb]{0.8674276350862745,0.864376599772549,0.8626024620196079}{\textbf{a}} \textcolor[rgb]{0.8674276350862745,0.864376599772549,0.8626024620196079}{\textbf{c}} \textcolor[rgb]{0.8674276350862745,0.864376599772549,0.8626024620196079}{\textbf{c}} \textcolor[rgb]{0.705673158,0.01555616,0.150232812}{\textbf{b}}  & \textcolor[rgb]{0.705673158,0.01555616,0.150232812}{\textbf{a}} \textcolor[rgb]{0.8674276350862745,0.864376599772549,0.8626024620196079}{\textbf{c}} \textcolor[rgb]{0.8674276350862745,0.864376599772549,0.8626024620196079}{\textbf{c}} \textcolor[rgb]{0.9648353582352941,0.7446136745882352,0.6432388753529412}{\textbf{b}}  \\
		4 & 0.999 & 0.761 &  \textcolor[rgb]{0.8674276350862745,0.864376599772549,0.8626024620196079}{\textbf{a}} \textcolor[rgb]{0.9648353582352941,0.7446136745882352,0.6432388753529412}{\textbf{c}} \textcolor[rgb]{0.705673158,0.01555616,0.150232812}{\textbf{c}} \textcolor[rgb]{0.8883904907411765,0.41770291749019606,0.32789791100392157}{\textbf{b}}   & \textcolor[rgb]{0.8674276350862745,0.864376599772549,0.8626024620196079}{\textbf{a}} \textcolor[rgb]{0.8674276350862745,0.864376599772549,0.8626024620196079}{\textbf{c}} \textcolor[rgb]{0.8674276350862745,0.864376599772549,0.8626024620196079}{\textbf{c}} \textcolor[rgb]{0.705673158,0.01555616,0.150232812}{\textbf{b}}  & \textcolor[rgb]{0.705673158,0.01555616,0.150232812}{\textbf{a}} \textcolor[rgb]{0.8674276350862745,0.864376599772549,0.8626024620196079}{\textbf{c}} \textcolor[rgb]{0.8674276350862745,0.864376599772549,0.8626024620196079}{\textbf{c}} \textcolor[rgb]{0.9593847296274509,0.6103057604117648,0.4893818509411765}{\textbf{b}}  \\
		8 & 1.000 & 0.762 &  \textcolor[rgb]{0.8674276350862745,0.864376599772549,0.8626024620196079}{\textbf{a}} \textcolor[rgb]{0.9648353582352941,0.7446136745882352,0.6432388753529412}{\textbf{c}} \textcolor[rgb]{0.705673158,0.01555616,0.150232812}{\textbf{c}} \textcolor[rgb]{0.8883904907411765,0.41770291749019606,0.32789791100392157}{\textbf{b}}   & \textcolor[rgb]{0.8674276350862745,0.864376599772549,0.8626024620196079}{\textbf{a}} \textcolor[rgb]{0.8674276350862745,0.864376599772549,0.8626024620196079}{\textbf{c}} \textcolor[rgb]{0.8674276350862745,0.864376599772549,0.8626024620196079}{\textbf{c}} \textcolor[rgb]{0.8674276350862745,0.864376599772549,0.8626024620196079}{\textbf{b}}  & \textcolor[rgb]{0.705673158,0.01555616,0.150232812}{\textbf{a}} \textcolor[rgb]{0.8674276350862745,0.864376599772549,0.8626024620196079}{\textbf{c}} \textcolor[rgb]{0.8674276350862745,0.864376599772549,0.8626024620196079}{\textbf{c}} \textcolor[rgb]{0.9057834780117647,0.4551856921647059,0.35533588384705883}{\textbf{b}}  \\
		16 & 1.000 & 0.762 & \textcolor[rgb]{0.8674276350862745,0.864376599772549,0.8626024620196079}{\textbf{a}} \textcolor[rgb]{0.9648353582352941,0.7446136745882352,0.6432388753529412}{\textbf{c}} \textcolor[rgb]{0.705673158,0.01555616,0.150232812}{\textbf{c}} \textcolor[rgb]{0.8883904907411765,0.41770291749019606,0.32789791100392157}{\textbf{b}}   & \textcolor[rgb]{0.8674276350862745,0.864376599772549,0.8626024620196079}{\textbf{a}} \textcolor[rgb]{0.8674276350862745,0.864376599772549,0.8626024620196079}{\textbf{c}} \textcolor[rgb]{0.8674276350862745,0.864376599772549,0.8626024620196079}{\textbf{c}} \textcolor[rgb]{0.8674276350862745,0.864376599772549,0.8626024620196079}{\textbf{b}}  & \textcolor[rgb]{0.9688625003647059,0.7108384838117647,0.5999005044313725}{\textbf{a}} \textcolor[rgb]{0.8674276350862745,0.864376599772549,0.8626024620196079}{\textbf{c}} \textcolor[rgb]{0.8674276350862745,0.864376599772549,0.8626024620196079}{\textbf{c}} \textcolor[rgb]{0.705673158,0.01555616,0.150232812}{\textbf{b}}  \\
		64 & 1.000 & 0.762 &  \textcolor[rgb]{0.8674276350862745,0.864376599772549,0.8626024620196079}{\textbf{a}} \textcolor[rgb]{0.9648353582352941,0.7446136745882352,0.6432388753529412}{\textbf{c}} \textcolor[rgb]{0.705673158,0.01555616,0.150232812}{\textbf{c}} \textcolor[rgb]{0.8883904907411765,0.41770291749019606,0.32789791100392157}{\textbf{b}}   & \textcolor[rgb]{0.8674276350862745,0.864376599772549,0.8626024620196079}{\textbf{a}} \textcolor[rgb]{0.8674276350862745,0.864376599772549,0.8626024620196079}{\textbf{c}} \textcolor[rgb]{0.8674276350862745,0.864376599772549,0.8626024620196079}{\textbf{c}} \textcolor[rgb]{0.8674276350862745,0.864376599772549,0.8626024620196079}{\textbf{b}}  & \textcolor[rgb]{0.8674276350862745,0.864376599772549,0.8626024620196079}{\textbf{a}} \textcolor[rgb]{0.8674276350862745,0.864376599772549,0.8626024620196079}{\textbf{c}} \textcolor[rgb]{0.8674276350862745,0.864376599772549,0.8626024620196079}{\textbf{c}} \textcolor[rgb]{0.705673158,0.01555616,0.150232812}{\textbf{b}}   \\\hline
	\end{tabular}
	\caption{Gradient-based heatmaps of $R^{(\textit{True})}_t(\ttt{accb})$ for the counter-based SP network, with $0.6 \leq u \leq 64$.}
	\label{fig:saturation}
\end{table}

\begin{table}
	\centering 
	\small
	\newcommand{\Sci}[2]{$-\text{#1} \times \text{10}^{-\text{#2}}$}
	\begin{tabular}{c c c c c}
		\hline  
		$m$ & $\sigma(m)$ & $c^{(t)}$ & \textbf{Accuracy} & \textbf{\% Blank} \\\hline
		4 & 0.982 & \Sci{8.74}{3} & 90.1 & 0.2 \\
		5 & 0.993 & \Sci{3.48}{3} & 96.1 & 2.2 \\
		6 & 0.998 & \Sci{1.32}{3} & 99.8 & 6.5 \\
		7 & 0.999 & \Sci{4.91}{4} & 100.0 & 22.0 \\
		8 & 1.000 & \Sci{1.81}{4} & 100.0 & 42.1 \\
		9 & 1.000 & \Sci{6.68}{5} & 100.0 & 69.9 \\
		10 & 1.000 & \Sci{2.46}{5} & 100.0 & 92.3 \\
		11 & 1.000 & \Sci{9.05}{6} & 100.0 & 98.7 \\
		12 & 1.000 & \Sci{3.33}{6} & 100.0 & 99.8 \\\hline
	\end{tabular}
	\caption{The results of the LRP saturation test, including the value of $m$, the average value of $c^{(t)}$ when the counter reaches $0$, the network's testing accuracy, and the percentage of examples with blank heatmaps for prefixes with equal numbers of \ta s and \tb s.}
	\label{fig:lrpzeros}
\end{table}

As mentioned in the previous section, network saturation causes gradients to be approximately $0$ when using sigmoid or $\tanh$ activation functions. To test how attribution methods are affected by saturation, \autoref{fig:saturation} shows heatmaps for the input \ttt{accb} generated by gradient-based methods for different instantiations of the counter-based SP network with varying degrees of saturation. Recall from \autoref{sec:whiteboxnetworks} that counter values for this network are expressed in multiples of the scaling factor $v$. We control the saturation of the network via the parameter $u = \tanh^{-1}(v)$. For all three gradient-based methods, scores for \ta\ decrease and scores for \tb\ increase as $u$ increases. Additionally, saliency scores for the first \tc\ decrease when $u$ increases. When $u = 8$, $v$ is almost completely saturated, causing G $\times$ I to produce all-zero heatmaps. On the other hand, IG is still able to produce nonzero heatmaps even at $u = 64$. Thus, IG is much more resistant to the effects of saturation than G $\times$ I. 

According to \citet{sundararajanAxiomaticAttributionDeep2017}, gradient-based methods satisfy the axiom of \textit{implementation invariance}: they produce the same heatmaps for any two networks that compute the same function. This formal property is seemingly at odds with the diverse array of heatmaps appearing in \autoref{fig:saturation}, which are produced for networks that all yield identical classifiers. In particular, the networks with $u = 8$, $16$, and $64$ yield qualitatively different heatmaps, despite the fact that the three networks are distinguished only by differences in $v$ of less than $0.001$. Because the three functions are technically not equal, implementation invariance is not violated in theory; but the fact that IG produces different heatmaps for three nearly identical networks shows that the intuition described by implementation invariance is not borne out in practice.

Besides the gradient-based methods, LRP is also susceptible to problems arising from saturation. Recall from heatmaps \#3 and \#4 of \autoref{fig:heatmaps} that for the counting task network, LRP assigns scores of $0$ to prefixes with equal numbers of \ta s and \tb s. We hypothesize that this phenomenon is related to the fact $c^{(t)} = 0$ after reading such prefixes, since the counter has been incremented and decremented in equal amounts. Accordingly, we test whether this phenomenon can be mitigated by desaturating the gates so that $c^{(t)}$ does not exactly reach $0$. Recall that the white-box LSTM gates approximate $1 \approx \sigma(m)$ using a constant $m \gg 0$. We construct networks with varying values of $m$ and compute LRP scores on a randomly generated testing set of 1000 strings, each of which contains at least one prefix with equal numbers of \ta s and \tb s. In \autoref{fig:lrpzeros} we report the percentage of examples for which such prefixes receive LRP scores of $0$, along with the network's accuracy on this testing set and the average value of $c^{(t)}$ when the counter reaches 0. Indeed, the percentage of prefixes receiving scores of $0$ increases as the approximation $c^{(t)} \approx 0$ becomes more exact.

\subsection{Ablation Test}
\label{sec:ablation}

\begin{table}
	\small 
	\centering
	\begin{tabular}{l c c}
		\hline
		\textbf{Method} & \textbf{SP (Counter)} & \textbf{SP (FSA)} \\\hline
		Occlusion & 61.8$_{\pm \text{12.2}}$ & \textbf{52.6}$_{\pm \text{11.7}}$ \\
		Saliency & 97.8$_{\pm \text{1.1}}$ & 96.0$_{\pm \text{2.5}}$ \\
		G $\times$ I & 65.7$_{\pm \text{14.4}}$ & 96.0$_{\pm \text{2.5}}$ \\
		IG & \textbf{47.5}$_{\pm \text{7.6}}$ & 94.9$_{\pm \text{2.9}}$ \\
		LRP & 64.3$_{\pm \text{12.7}}$ & 96.0$_{\pm \text{2.5}}$ \\\hline
		Random & \multicolumn{2}{c}{96.1$_{\pm \text{2.5}}$} \\
		Optimal & \multicolumn{2}{c}{\textbf{42.7}$_{\pm \text{3.8}}$} \\\hline
	\end{tabular}
	\caption{Mean and standard deviation results of the ablation test, normalized by string length and expressed as a percentage. ``Optimal'' is the best possible score.}
	\label{fig:ablation}
\end{table}

So far, we have primarily compared attribution methods via visual inspection of individual examples. To compare the five methods quantitatively, we apply the ablation test of \citet{bachPixelWiseExplanationsNonLinear2015} to our two white-box networks for the SP task.\footnote{We do not consider the counting task because its heatmaps are already easy to understand, and we do not consider the PDA network because the gradient-based methods fail to produce nonzero heatmaps for that network.} Given an input string classified as \textit{True}, we iteratively remove the symbol with the highest relevance score, recomputing heatmaps at each iteration, until the string no longer contains any of the four subsequences. We apply the ablation test to 100 randomly generated input strings, and report the average percentage of each string that is ablated in \autoref{fig:ablation}. A peculiar property of the SP task is that removing a symbol preserves the validity of input strings. This means that, unlike in NLP settings, our ablation test does not suffer from the issue that ablation produces invalid inputs.

Saliency, G $\times$ I, and LRP perform close to the random baseline on the FSA network; this is unsurprising, since these methods only assign nonzero scores to the last input symbol. While \autoref{fig:heatmaps} shows some variation in the IG heatmaps, IG also performs close to the random baseline. Only occlusion performs considerably better, since it is able to identify symbols whose ablation would destroy subsequences.

On the counter-based SP network, IG performs remarkably close to the optimal benchmark, which represents the best possible performance on this task. Occlusion, G $\times$ I, and LRP achieve a similar level of performance to one another, while saliency performs worse than the random baseline.

\section{Conclusion}
\label{sec:conclusion}

Of all the heatmaps considered in this paper, only those computed by G $\times$ I and IG for the counting task fully matched our expectations. In other cases, all attribution methods fail to identify at least some of the input features that should be considered relevant, or assign relevance to input features that do not affect the model's behavior. Among the five methods, saliency achieves the worst performance: it never assigns nonzero scores for the counting and bracket prediction tasks, and it does not identify the relevant symbols for either of the two SP networks. Saliency also achieves the worst performance on the ablation test for both the counter-based and the FSA-based SP networks. Among the four white-box networks, the two automata-based networks proved to be much more challenging for the attribution methods than the counter-based networks. While the LRP heatmaps for the PDA network correctly identify the matching bracket when available, no other method produces reasonable heatmaps for the PDA network, and all five methods fail to interpret the FSA network.

Taken together, our results suggest that attribution heatmaps should be viewed with skepticism. This paper has identified cases in which heatmaps fail to highlight relevant features, as well as cases in which heatmaps incorrectly highlight irrelevant features. Although most of the methods perform better for the counter-based networks than the automaton-based networks, in practical settings we do not know what kinds of computations are implemented by a trained network, making it impossible to determine whether the network under analysis is compatible with the attribution method being used.

In future work, we encourage the use of our four white-box models as qualitative benchmarks for evaluating interpretability methods. For example, the style of evaluation we have developed can be replicated for attribution methods not covered in this paper, including DeepLIFT \citep{shrikumarNotJustBlack2017} and contextual decomposition \citep{murdochWordImportanceContextual2018}. We believe that insights gleaned from white-box analysis can help researchers choose between different attribution methods and identify areas of improvement in current techniques.

\section*{Acknowledgments}

I would like to thank Dana Angluin and Robert Frank for their advice and mentorship on this project. I would also like to thank Yoav Goldberg, John Lafferty, Tal Linzen, R. Thomas McCoy, Aaron Mueller, Karl Mulligan, Shauli Ravfogel, Jason Shaw, and the reviewers for their helpful feedback and discussion.

\bibliography{bibliography}
\bibliographystyle{acl_natbib}

\appendix
\section{Detailed Descriptions of White-Box Networks}
\label{sec:appendixa}

This appendix provides detailed descriptions of our four white-box networks.

\subsection{Counting Task Network}

As described in \autoref{sec:counternetworks}, the network for the counting task simply sets $g^{(t)}$ to $v = \tanh(u)$ when $\bm{x}^{(t)} = \ta$ and $-v$ when $\bm{x}^{(t)} = \tb$. All gates are fixed to $1$. The output layer uses $h^{(t)} = \tanh\left(c^{(t)}\right)$ as the score for the \textit{True} class and $v/2$ as the score for the \textit{False} class.
\begin{align*}
g^{(t)} &= \tanh\left(u \left[
\begin{array}{c c}
1 & -1
\end{array}
\right] \bm{x}^{(t)} \right) \\
f^{(t)} &= \sigma(m) \\
i^{(t)} &= \sigma(m) \\
o^{(t)} &= \sigma(m) \\
\hat{\bm{y}}^{(t)} &= \left[\begin{array}{c c}
1 \\ 0
\end{array}\right] h^{(t)} + \left[\begin{array}{c c}
0 \\ v/2
\end{array}\right]
\end{align*}

\subsection{SP Task Network (Counter-Based)}

The seven counters for the SP task are implemented as follows. First, we compute $\bm{g}^{(t)}$ under the assumption that one of the first four counters is always incremented, and one of the last three counters is always incremented as long as $\bm{x}^{(t)} \neq \ta$.
\[
\bm{g}^{(t)} = \tanh\left( u
\left[\begin{array}{c}
\bm{I}_4 \\\hdashline
\begin{array}{cccc}
0 & 1 & 0 & 0 \\
0 & 0 & 1 & 0 \\
0 & 0 & 0 & 1
\end{array}
\end{array}\right] \bm{x}^{(t)}\right)
\]
Then, we use the input gate to condition the last three counters on the value of the first four counters. For example, if $h_1^{(t - 1)} = 0$, then no \ta s have been encountered in the input string before time $t$. In that case, the input gate for counter \#5, which represents subsequences ending with \tb, is set to $i_5^{(t)} = \sigma(-m) \approx 0$. This is because a \tb\ encountered at time $t$ would not form part of a subsequence if no \ta s have been encountered so far, so counter \#5 should not be incremented.
\begin{align*}
\bm{i}^{(t)} &= \sigma\left( 2m \left[\begin{array}{c:c}
\bm{0} & \bm{0} \\\hdashline
\begin{array}{cccc}
1 & 0 & 0 & 0  \\
0 & 1 & 0 & 1  \\
0 & 0 & 1 & 0 
\end{array} &
\bm{0}
\end{array}\right] \bm{h}^{(t - 1)} \right. \\
&\mathrel{\phantom{=}} \left. + m \left[ \begin{array}{c c c c c c c}
1 & 1 & 1 & 1 & -1 & -1 & -1
\end{array}\right]^\top \right)
\end{align*}
All other gates are fixed to $\bm{1}$. The output layer sets the score of the \textit{True} class to $h_5^{(t)} + h_6^{(t)} + h_7^{(t)}$ and the score of the \textit{False} class to $v/2$.
\begin{align*}
\bm{f}^{(t)} &= \sigma(m\bm{1}) \\
\bm{o}^{(t)} &= \sigma(m\bm{1}) \\
\hat{\bm{y}}^{(t)} &= \left[\begin{array}{c : c c c}
\bm{0} & 1 & 1 & 1 \\ 
\bm{0} & 0 & 0 & 0
\end{array}\right] \bm{h}^{(t)} + \left[\begin{array}{c c}
0 \\ v/2
\end{array}\right]
\end{align*}

\subsection{FSA Network}
\label{sec:fsadetails}

Here we describe a general construction of an LSTM simulating an FSA with states $Q$, accepting states $Q_F \subseteq Q$, alphabet $\Sigma$, and transition function $\delta: Q \times \Sigma \to Q$. Recall that $\bm{h}^{(t)}$ contains a one-hot representation of pairs in $Q \times \Sigma$ encoding the current state of the FSA and the most recent input symbol. The initial state $\bm{h}^{(0)} = \bm{0}$ represents the starting configuration of the FSA.

At a high level, the state transition system works as follows. First, $\bm{g}^{(t)}$ first marks all the positions corresponding to the current input $\bm{x}^{(t)}$.\footnote{We use $v = \tanh(1) \approx 0.762$.}
\[
g^{(t)}_{\langle q, x \rangle} = \begin{cases}
v, & x = \bm{x}^{(t)} \\
0, & \text{otherwise}
\end{cases}
\]
The input gate then filters out any positions that do not represent valid transitions from the previous state $q^\prime$, which is recovered from $\bm{h}^{(t - 1)}$.
\[
i^{(t)}_{\langle q, x \rangle} = \begin{cases}
1, & \delta(q^\prime, x) = q \\
0, & \text{otherwise}
\end{cases}
\]
Now, we describe how this behavior is implemented in our LSTM.

The cell state update is straightforwardly implemented as follows:
\[
\bm{g}^{(t)} = \tanh\left( u\bm{W}^{(c, x)}\bm{x}^{(t)} \right)\text{,}
\]
where
\[
W^{(c, x)}_{\langle q, x \rangle, j} = \begin{cases}
1, & \text{$j$ is the index for $x$} \\
0, & \text{otherwise.}
\end{cases}
\]
Observe that the matrix $\bm{W}^{(c, x)}$ essentially contains a copy of $\bm{I}_4$ for each state, such that each copy is distributed across the different cell state units designated for that state. 

The input gate is more complex. First, the bias term handles the case where the current case is the starting state $q_0$. This is necessary because the initial configuration of the network is represented by $\bm{h}^{(0)} = \bm{0}$.
\[
b^{(i)}_{\langle q, x \rangle} = \begin{cases}
m, & \delta(q_0, x) = q \\
-m, & \text{otherwise}
\end{cases}
\]
The bias vector sets $i^{(t)}_{\langle q, x \rangle}$ to be $1$ if the FSA transitions from $q_0$ to $q$ after reading $x$, and $0$ otherwise. We replicate this behavior for other values of $\bm{h}^{(t - 1)}$ by using the weight matrix $\bm{W}^{(i, h)}$, taking the bias vector into account:
\[
\bm{i}^{(t)} = \sigma\left( \bm{W}^{(i, h)}\bm{h}^{(t - 1)} + \bm{b}^{(i)} \right)\text{,}
\] 
where
\[
W^{(i)}_{\langle q, x \rangle, \langle q^\prime, x^\prime \rangle} = \begin{cases}
m - b^{(i)}_{\langle q, x \rangle}, &  \delta(q^\prime, x) = q\\
-m - b^{(i)}_{\langle q, x \rangle}, & \text{otherwise.}
\end{cases}
\]
The forget gate is fixed to $-\bm{1}$, since the state needs to be updated at every time step. The output gate is fixed to $\bm{1}$.
\begin{align*}
\bm{f}^{(t)} &= \sigma(-m\bm{1}) \\
\bm{o}^{(t)} &= \sigma(m\bm{1})
\end{align*}
The output layer simply selects hidden units that represent accepting and rejecting states:
\[
\hat{\bm{y}}^{(t)} = \bm{W}\bm{h}^{(t)}\text{,}
\]
where
\[
W_{c, \langle q, x \rangle} = \begin{cases}
1, & c = \textit{True} \mathrel{\text{and}} q \in Q_F \\
1, & c = \textit{False} \mathrel{\text{and}} q \notin Q_F \\
0, & \text{otherwise.}
\end{cases}
\]

\subsection{PDA Network}
\label{sec:pdaappendix}

Finally, we describe how the PDA network for the bracket prediction task is implemented. Of the four networks, this one is the most intricate. Recall from \autoref{sec:automatanetworks} that we implement a bounded stack of size $k$ using $2k + 1$ hidden units, with the following interpretation:
\begin{itemize}
	\item $\bm{c}^{(t)}_{:k - 1}$ contains the stack, except for the top item \\\
	\item $c^{(t)}_k$ contains the top item of the stack
	\item $\bm{c}^{(t)}_{k + 1:2k}$ contains the height of the stack in unary notation \\
	\item $c_{2k + 1}$ is a bit, which is set to be positive if the stack is empty and nonpositive otherwise.
\end{itemize}
We represent the brackets \ttt{(}, \ttt{[}, \ttt{)}, and \ttt{]} in one-hot encoding with the indices $1$, $2$, $3$, and $4$, respectively. The opening brackets \ttt{(} and \ttt{[} are represented on the stack by $1$ and $-1$, respectively. T

We begin by describing $\bm{g}^{(t)}$. Due to the complexity of the network, we describe the weights and biases individually, which are combined as follows.
\begin{align*}
\bm{g}^{(t)} &= \tanh\left(m\left(\bm{z}^{(g, t)} \right)\right)\text{, where} \\
\bm{z}^{(g, t)} &= \bm{W}^{(c, x)}\bm{x}^{(t)} + \bm{W}^{(c, h)}\bm{h}^{(t - 1)} + \bm{b}^{(c)}
\end{align*}
First, the bias vector sets $c^{(t)}_{2k + 1}$ to be $1$, indicating that the stack is empty. This ensures that the initial hidden state $\bm{h}^{(t)} = \bm{0}$ is treated as an empty stack.
\[
\bm{b}^{(c)} = \left[ \begin{array}{c}
\bm{0} \\ \hdashline
2
\end{array} \right]
\]
$\bm{W}^{(c, x)}$ serves three functions when $\bm{x}^{(t)}$ is an open bracket, and does nothing when $\bm{x}^{(t)}$ is a closing bracket. First, it pushes $\bm{x}^{(t)}$ to the top of the stack, represented by $c^{(t)}_k$. The values $\pm 2^k$ are determined by equation (1) in Subsection 4.2. Second, it sets $\bm{g}^{(t)}_{k + 1:2k}$ to $\bm{1}$ in order to increment the unary counter for the height of the stack. Later, we will see that the input gate filters out all positions except for the top of the stack. Finally, $\bm{W}^{(c, x)}$ sets the empty stack indicator to $-1$, indicating that the stack is not empty.
\[
\bm{W}^{(c, x)} = \left[ \begin{array}{c c c c}
\bm{0} & \bm{0} & \bm{0} & \bm{0} \\\hdashline
2^k & -2^k & 0 & 0 \\ \hdashline
\bm{1} & \bm{1} & \bm{0} & \bm{0} \\\hdashline
-2 & -2 & 0 & 0 
\end{array} \right]
\]
$\bm{W}^{(c, h)}$ performs two functions. First, it completes equation (1) for $c_{k}^{(t)}$, setting it to be the second-highest stack item from the previous time step. Second, it copies the top of the stack to the first $k - 1$ positions, with the input gate filtering out all but the highest position.
\[
\bm{W}^{(c, h)} = \left[ \begin{array}{c : c : c : c}
\bm{0} & \bm{1} & \bm{0} & \bm{0} \\\hdashline
\begin{array}{cccc}
2 & 4 & \cdots & 2^{k - 1} 
\end{array} & 0 & \bm{0} & 0\\\hdashline
\bm{0} & \bm{0} & \bm{0} & \bm{0} \\\hdashline
\bm{0} & 0 & -\bm{1} & 0
\end{array} \right]
\]
Finally, the $-1$s serve to decrease the empty stack indicator by an amount proportional to the stack height at time $t - 1$.  Observe that if $\bm{x}^{(t)}$ is a closing bracket and $\bm{h}^{(t - 1)}$ represents a stack with only one item, then 
\begin{align*}
&\mathrel{\phantom{=}} \bm{W}^{(c, x)}_{2k + 1, :}\bm{x}^{(t)} + \bm{W}^{(c, h)}_{2k + 1, :}\bm{h}^{(t - 1)} + b^{(c)}_{2k + 1} \\
&= -1 + 2 = 1\text{,}
\end{align*}
so the empty stack indicator is set to $1$, indicating that the stack is empty. Otherwise,
\[
\bm{W}^{(c, x)}_{2k + 1, :}\bm{x}^{(t)} + \bm{W}^{(c, h)}_{2k + 1, :}\bm{h}^{(t - 1)} \leq -2\text{,}
\]
so the empty stack indicator is nonpositive.

Now, we describe the input gate, given by the following.
\begin{align*}
\bm{i}^{(t)} &= \sigma\left(m\left(\bm{z}^{(i, t)} \right)\right) \\
\bm{z}^{(i, t)} &= \bm{W}^{(i, x)}\bm{x}^{(t)} + \bm{W}^{(i, h)}\bm{h}^{(t - 1)} + \bm{b}^{(i)}
\end{align*}
$\bm{W}^{(i, x)}$ sets the input gate for the first $k - 1$ positions to $0$ when $\bm{x}^{(t)}$ is a closing bracket. In that case, an item needs to be popped from the stack, so nothing can be copied to these hidden units. When $\bm{x}^{(t)}$ is an opening bracket, $\bm{W}^{(i, x)}$ sets $i_k^{(t)} = 1$, so that the bracket can be copied to the top of the stack.
\[
\bm{W}^{(i, x)} = 2\left[
\begin{array}{c}
\begin{array}{c c c c}
\bm{0} &  \bm{0} & -\bm{1} & -\bm{1} \\\hdashline

1 & 1 & 0 & 0 
\end{array} \\\hdashline 
\bm{0}
\end{array}
\right]
\]
$\bm{W}^{(i, h)}$ uses a matrix $\bm{T}_n \in \mathbb{R}^{n \times n}$, defined below.
\[
\bm{T}_n = \left[\begin{array}{c c c c c c}
1 & -1 & 0 & \dots & 0 & 0 \\
0 & 1 & -1 & \dots & 0 & 0 \\
\vdots & \vdots & \vdots & \ddots & \vdots & \vdots \\
0 & 0 & 0 & \dots & 1 & -1 \\
0 & 0 & 0 & \dots & 0 & 1
\end{array}\right]
\]
Suppose $\bm{v}$ represents the number $s$ in unary notation: $v_j$ is $1$ if $j \leq s$ and $0$ otherwise. $\bm{T}_n$ has the special property that $\bm{T}_n\bm{v}$ is a one-hot vector for $s$. Based on this, $\bm{W}^{(i, h)}$ is defined as follows.
\[
\bm{W}^{(i, h)} = 2\left[ 
\begin{array}{c:c:c}
\bm{0} & \begin{array}{c}
\left( \bm{T}_k \right)_{:k - 1, :} \\\hdashline
\bm{0} \\\hdashline
\left( \bm{T}_k \right)_{:k - 1, :} \\\hdashline
\bm{0}
\end{array}& \bm{0}
\end{array}
\right]
\]
$\bm{W}^{(i, h)}_{:k - 1, k + 1:2k}$ contains $\bm{T}_k$, with the last row truncated. This portion of the matrix converts $\bm{h}^{(t - 1)}_{k + 1:2k}$, which contains a unary encoding of the stack height, to a one-hot vector marking the position of the top of the stack. This ensures that, when pushing to the stack, the top stack item from time $t - 1$ is only copied to the appropriate position of $\bm{h}_{:k - 1}^{(t)}$. The other copy of $\bm{T}_k$, again with the last row omitted, occurs in $\bm{W}^{(i, h)}_{k + 2:2k, k + 1:2k}$. This copy of $\bm{T}_k$ ensures that when the unary counter for the stack height is incremented, only the appropriate position is updated. Finally, the bias vector ensures that the top stack item and the empty stack indicator are always updated.
\[
\bm{b}^{(i)} = \left[ \begin{array}{c}
-\bm{1} \\ \hdashline
1 \\ \hdashline 
-\bm{1} \\\hdashline
1
\end{array} \right]
\]

The forget gate is responsible for deleting portions of memory when stack items are popped.
\begin{align*}
\bm{f}^{(t)} &= \sigma\left(m\left( \bm{z}^{(f, t)} \right)\right) \\
\bm{z}^{(f, t)} &= \bm{W}^{(f, x)}\bm{x}^{(t)} + \bm{W}^{(f, h)}\bm{h}^{(t - 1)} + \bm{b}^{(f)}
\end{align*}
$\bm{W}^{(f, x)}$ first ensures that no stack items are deleted when an item is pushed to the stack.
\[
\bm{W}^{(f, x)} = 2\left[
\begin{array}{c c c c}
\bm{1} & \bm{1} & \bm{0} & \bm{0} \\\hdashline
0 & 0 & 0 & 0 \\\hdashline
\bm{1} & \bm{1} & \bm{0} & \bm{0} \\\hdashline
0 & 0 & 0 & 0 
\end{array}
\right]
\]
Next, $\bm{W}^{(f, h)}$ marks the second highest stack position and the top of the unary counter for deletion, in case an item needs to be popped.
\[
\bm{W}^{(f, h)} = 2\left[ 
\begin{array}{c:c:c}
\bm{0} & \begin{array}{c}
-\left( \bm{T}_k \right)_{2:, :} \\\hdashline
\bm{0} \\\hdashline
-\bm{T}_k  \\\hdashline
\bm{0}
\end{array}& \bm{0}
\end{array}
\right]
\]
Finally, the bias term ensures that the top stack item and empty stack indicator are always cleared.
\[
\bm{b}^{(i)} = \left[ \begin{array}{c}
\bm{1} \\ \hdashline
-1 \\ \hdashline 
\bm{1} \\\hdashline
-1
\end{array} \right]
\]

To complete the construction, we fix the output gate to $\bm{1}$, and have the output layer read the top stack position:
\begin{align*}
\bm{o}^{(t)} &= \sigma(m\bm{1}) \\
\hat{\bm{y}}^{(t)} &= \bm{W}\bm{h}^{(t)}\text{,}
\end{align*}
where
\[
W_{c, j} = \begin{cases}
1, & c = \ttt{)} \mathrel{\text{and}} j = k \\
-1, & c = \ttt{]} \mathrel{\text{and}} j = k \\
1, & c = \textit{None} \mathrel{\text{and}} j = 2k + 1 \\
0, & \text{otherwise.}
\end{cases}
\]

\end{document}